\documentclass[runningheads]{llncs}
\usepackage{graphicx}
\usepackage{amsmath,amssymb} 

\usepackage[width=122mm,left=12mm,paperwidth=146mm,height=193mm,top=12mm,paperheight=217mm]{geometry}

\usepackage{color}
\usepackage[capitalise]{cleveref} 
\usepackage{booktabs}
\usepackage{array}
\usepackage{algorithm}
\usepackage{algpseudocode}
\usepackage{algorithmicx}

\begin{document}
\pagestyle{headings}
\mainmatter

\def\ACCV22SubNumber{357}  

\title{Physical Passive Patch Adversarial Attacks on Visual Odometry Systems} 

\authorrunning{Yaniv Nemcovsky et al.} 
\author{Yaniv Nemcovsky$^{\dagger}{}^{1}$ \and Matan Jacoby $^{\dagger}{}^{1}$ \and  Alex M. Bronstein$^{1}$ \and Chaim Baskin$^{1}$ }

\institute{$^{1}$ Department of Computer Science -- Technion, Haifa, Israel\\
$^{\dagger}$ equal contribution}



\maketitle

\begin{abstract}
Deep neural networks are known to be susceptible to adversarial perturbations -- small perturbations that alter the output of the network and exist under strict norm limitations. While such perturbations are usually discussed as tailored to a specific input, a universal perturbation can be constructed to alter the model's output on a set of inputs. Universal perturbations present a more realistic case of adversarial attacks, as awareness of the model’s exact input is not required. In addition, the universal attack setting raises the subject of generalization to unseen data, where given a set of inputs, the universal perturbations aim to alter the model’s output on out-of-sample data. In this work, we study physical passive patch adversarial attacks on visual odometry-based autonomous navigation systems. A visual odometry system aims to infer the relative camera motion between two corresponding viewpoints, and is frequently used by vision-based autonomous navigation systems to estimate their state. For such navigation systems, a patch adversarial perturbation poses a severe security issue, as it can be used to mislead a system onto some collision course. To the best of our knowledge, we show for the first time that the error margin of a visual odometry model can be significantly increased by deploying patch adversarial attacks in the scene. We provide evaluation on synthetic closed-loop drone navigation data and demonstrate that a comparable vulnerability exists in real data.
A reference implementation of the proposed method and the reported experiments is provided at \url{https://github.com/patchadversarialattacks/patchadversarialattacks}. 
\end{abstract}

\section{Introduction}
\label{sec:intro}

\paragraph{Adversarial attacks.}

Deep neural networks (DNNs) were the first family of models discovered to be susceptible to adversarial perturbations --  small bounded-norm perturbations of the input that significantly alter the output of the model \cite{szegedy2013intriguing,goodfellow2014explaining} (methods for producing such perturbations are referred to as adversarial attacks). Such perturbations are usually discussed as tailored to a specific model and input, and in such settings were shown to undermine the impressive performance of DNNs across multiple fields, e.g., object detection  \cite{wang2019daedalus}, real-life object recognition \cite{brown2017adversarial,xu2019evading,athalye2017synthesizing}, reinforcement learning \cite{Gleave2020Adversarial}, speech-to-text \cite{carlini2018audio}, point cloud classification \cite{xiang2019generating}, natural language processing \cite{gao2018black,chaturvedi2019exploring,jin2020bert}, video recognition \cite{pony2021over}, Siamese Visual Tracking \cite{li2021simple}, and on several regression tasks \cite{nguyen2018adversarial,mode2020adversarial,yamanaka2020adversarial} as well as autonomous driving \cite{deng2020analysis}. Moreover, adversarial attacks were shown to be transferable between models; i.e., an adversarial perturbation that is effective on one model will likely be effective for other models as well \cite{szegedy2013intriguing}. Recent studies also suggest that the vulnerability is a property of high-dimensional input spaces rather than of specific model classes \cite{gilmer2018adversarial,dube2018high,amsaleg2020high}.

Universal adversarial attacks are another setting where the aim is to produce an adversarial perturbation for a set of inputs \cite{moosavi2017universal,hendrik2017universal,zhang2021survey}. Universal perturbations present a more realistic case of adversarial attacks, as awareness of the model’s exact input is not required. In addition, the universal attack setting raises the subject of generalization to unseen data, where given a set of inputs, the universal perturbations aim to alter the model's output on out-of-sample data. In this setting, universal perturbations can also be used to improve the performance of DNNs on out-of-sample data \cite{salman2021unadversarial}.

\paragraph{Adversarial attacks on visual odometers.}
Monocular visual odometry (VO) models aim to infer the relative camera motion (position and orientation) between two corresponding viewpoints. Recently, DNN-based VO models have outperformed traditional monocular VO methods \cite{bian2019unsupervised,yang2020d3vo,almalioglu2022selfvio,wang2020tartanvo}. Specifically, the model suggested by Wang et al. (2020b)\cite{wang2020tartanvo} shows a promising ability to generalize from simulated training data to real scenes. Such models usually make use of either feature matching or photometric error minimization to compute the camera motion.

Vision-based autonomous navigation systems frequently use VO models as a method of estimating their state. Such navigation systems would use the trajectory estimated by the VO to compute their heading, closing the loop with the navigation control system that directs the vehicle to a target position in the scene. Visual simultaneous localization and mapping (visual SLAM, or vSLAM for short) techniques also make use of VO models to estimate the vehicle trajectory, additionally estimating the environment map and thereby adding global consistency to the estimations \cite{macario2022comprehensive,pinkovich2020predictive,mur2015orb,triggs1999bundle}. Adversarial attacks on VO models, consequently, pose a severe security issue for visual SLAM, as they could corrupt the estimated map and mislead the navigation. 
A recent work \cite{chawla2022adversarial} had discussed adversarial attacks on the monocular VO model over single image pairs, and shows the susceptibility of the estimated position and orientation to adversarial perturbations.

In the present work, we investigate the susceptibility of VO models to universal adversarial perturbations over trajectories with multiple viewpoints, aiming to mislead a corresponding navigation system by disrupting its ability to spatially position itself in the scene. Previous works that discuss adversarial attacks on regression models mostly discuss standard adversarial attacks where the perturbation is inserted directly into a single image
\cite{nguyen2018adversarial,mode2020adversarial,deng2020analysis,yamanaka2020adversarial,li2021simple,chawla2022adversarial}. In contrast, we take into consideration a time evolving process where a physical passive patch adversarial attack is inserted into the scene and is perceived differently from multiple viewpoints. This is a highly realistic settings, as we test the effect of a moving camera in a perturbed scene, and do not require direct access to the model's input. Below, we outline our main contributions.


Firstly, we produce physical patch adversarial perturbations for VO systems on both synthetic and real data. Our experiments show that while VO systems are robust to random perturbations, they are susceptible to such adversarial perturbations. For a given trajectory containing multiple frames, our attacks are aimed to maximize the generated deviation in the physical translation between the accumulated trajectory motion estimated by the VO and the ground truth. We show that inserting a physical passive adversarial patch into the scene substantially increases the generated deviation.

Secondly, we continue to produce universal physical patch adversarial attacks, which are aimed at perturbing unseen data. We optimize a single adversarial patch on multiple trajectories and test the attack on out-of-sample unseen data. Our experiments show that when used on out-of-sample data, our universal attacks generalize and again cause significant deviations in trajectory estimates produced by the VO system.

Lastly, we further test the robustness of VO systems to our previously produced universal adversarial attacks in a closed-loop scheme with a simple navigation scheme, on synthetic data. Our experiments show that in this case as well, the universal attacks force the VO system to deviate from the ground truth trajectory.
To the best of our knowledge, ours is the first time the vulnerability of visual navigation systems to adversarial attacks is demonstrated, and, possibly, the first instance of adversarial attacks on closed-loop control systems. 

The rest of the paper is organized as follows: \cref{sec:method} describes our proposed method, \cref{sec:exp} provides our experimental results, and \cref{sec:conclusion} concludes the paper.
\section{Method}
\label{sec:method}
Below, we start with a definition of the adversarial attack setting, for both the universal and standard cases. We then describe the adversarial optimization scheme used for producing the perturbations and discuss the optimization of the universal attacks aiming to perturb unseen data.

\subsection{Patch adversarial attack setting}
\paragraph{Patch adversarial perturbation.}
Let $\mathcal{I}=[0,1]^{3\times w\times h}$ be a normalized RGB image space, for some width $w$ and height $h$. For an image $I\in \mathcal{I}$, inserting a patch image $P\in \mathcal{I}$ onto a given plane in $I$ would then be a perturbation $A:(\mathcal{I}\times\mathcal{I}) \to \mathcal{I}$. We denote $I^P=A(I, P)$. To compute $I^P$, we first denote the black and white albedo images of the patch $P$ as viewed from viewpoint $I$, namely $I^0, I^1 \in \mathcal{I}$. The albedo images are identical except for pixels corresponding to $P$, which contains the minimal and maximal RGB intensity values of the patch from viewpoint $I$. As such, $I^0$ and $I^1$ essentially describe the dependency of $I^P$ on the lighting conditions and the material comprising $P$. In addition, we denote the linear homography transformation of $P$ to viewpoint $I$ as $H:\mathcal{I} \to \mathcal{I}$, such that $H(P)$ is the mapping of pixels from $P$ to $I$, without taking into consideration changes in the pixels' intensity. Note that $H$ is only dependent on the relative camera motion between $I$ and $P$. We now define $I^P$ as:
\begin{align}
    I^P = A(I, P) = H(P)*(I^1 - I^0) + I^0
\end{align}
where $*$ denotes element-wise multiplication. For a set of images $\{I_t\}$, we similarly define the perturbed set as inserting a single patch $P$ onto the same plane in each image:
\begin{align}
    \{I_t^P\} = A(\{I_t\}, P) = \{H_t(P)*(I_t^1 - I_t^0) + I_t^0\} \label{eq:adv_pert}
\end{align}

\paragraph{Attacking visual odometry}

Let $VO: (\mathcal{I}\times \mathcal{I}) \to (\mathbb{R}^3\times so(3))$ be a monocular VO model, i.e., for a given pair of consecutive images $\{I_t, I_{t+1}\}$, it estimates the relative camera motion $\delta_t^{t+1}=(q_t^{t+1}, R_t^{t+1})$, where $q_t^{t+1}\in \mathbb{R}^3$ is the $3D$ translation and $R_t^{t+1}\in so(3)$ is the $3D$ rotation. We define a trajectory as a set of consecutive images $\{I_t\}_{t=0}^L$, for some length $L$, and extend the definition of the monocular visual odometry to trajectories:
\begin{align}
    VO(\{I_t\}_{t=0}^L) = \{\hat{\delta}_t^{t+1}\}_{t=0}^{L-1}
\end{align}
where $\hat{\delta}_t^{t+1}$ denotes the estimation of $\delta_t^{t+1}$ by the VO model. Given a trajectory $\{I_t\}_{t=0}^L$, with ground truth motions $ \{\delta_t^{t+1}\}_{t=0}^{L-1}$ and a criterion over the trajectory motions $\ell$, an adversarial patch perturbation $P_a\in \mathcal{I}$ aims to maximize the criterion over the trajectory. Similarly, for a set of trajectories $\{\{I_{i,t}\}_{t=0}^{L_i}\}_{i=0}^{N-1}$, with corresponding ground truth motions $\{\{\delta_{i,t}^{t+1}\}_{t=0}^{{L_i}-1}\}_{i=0}^{N-1}$, a universal adversarial attack aims to maximize the sum of the criterion over the trajectories. Formally:
\begin{align}
    P_a &= \arg \max_{P \in \mathcal{I}} \ell(VO(A(\{I_t\}_{t=0}^L, P)), \{\delta_t^{t+1}\}_{t=0}^{L-1}) \\
    P_{ua} &= \arg \max_{P \in \mathcal{I}} \sum_{i=0}^{N-1} \ell(\{VO(A(I_{i,t}, P))\}_{t=0}^{L_i}, \{\delta_{i,t}^{t+1}\}_{t=0}^{{L_i}-1})
\end{align}
where $A$ is defined as in \cref{eq:adv_pert}, and the scope of the adversarial attacks' universality is according to the domain and data distribution of the given trajectories set. In this formulation, the limitation of the adversarial perturbation is expressed in the albedo images $I^0$ and $I^1$ and can be described as dependent on the patch material. In contrast to standard adversarial perturbations that are tailored to a specific input, we consider the generalization properties of universal adversarial perturbation to unseen data. In such cases the provided trajectories used for perturbation optimization would differ from the test trajectories.


\paragraph{Task criterion.}
For the scope of this paper, the target criterion used for adversarial attacks is the RMS (root mean square) deviation in the $3D$ physical translation between the accumulated trajectory motion, as estimated by the VO, and the ground truth. We denote the accumulated motion as $\delta_0^L = \delta_0^1 \cdot \delta_1^2 \cdots \delta_{L-1}^{L} = \prod_{t=0}^{L-1} \delta_t^{t+1}$, where the multiplication of motions is defined as the matrix multiplication of the corresponding $4\times4$ matrix representation: $\delta_t^{t+1} = \begin{pmatrix}
                          R_t^{t+1} & q_t^{t+1} \\
                          \textbf{0} & 1
                        \end{pmatrix}$.
                        
The target criterion is then formulated as:

\begin{align}
    \ell_{VO} (VO(A(\{I_t\}_{t=0}^L, P)), \{\delta_t^{t+1}\}_{t=0}^{L-1})
     &= ||q(\prod_{t=0}^{L-1} VO(I_t^P, I_{t+1}^P)) - q(\prod_{t=0}^{L-1} \delta_t^{t+1})||_2 
\end{align}
where we denote $q(\delta_0^L)=q((q_0^L, R_0^L))=q_0^L$.

\subsection{Optimization of adversarial patches}

We optimize the adversarial patch $P$ via a PGD adversarial attack \cite{madry2017towards} with $\ell_{inf}$ norm limitation. We limit the values in $P$ to be in $[0, 1]$; however, we do not enforce any additional $\epsilon$ limitation, as such would be expressed in the albedo images.
We allow for different training and evaluation criteria in both attack types, and to enable evaluation on unseen data for universal attacks, we allow for different training and evaluation datasets. In the supplementary material, we provide algorithms for both our PGD (\cref{alg:attack}) and universal (\cref{alg:universal_attack}) attacks. We note that the PGD attack is a specific case of the universal in which both training and evaluation datasets comprise the same single trajectory.


\paragraph{Optimization and evaluation criteria.}

For both optimization and evaluation of attacks we consider one of two criteria. The first criterion, which we denote as $\ell_{RMS}$, is a smoother version of the target criterion $\ell_{VO}$, in which we sum over partial trajectories with the same origin as the full trajectory. Similarly, the second criterion, which we denote as $\ell_{MPRMS}$, i.e., mean partial RMS, differ from $\ell_{RMS}$ by taking into account all the partial trajectories. Nevertheless, we take the mean for each length of partial trajectories in order to keep the factoring between different lengths as in $\ell_{RMS}$. $\ell_{MPRMS}$ may be more suited to generalization of universal attacks to unseen data than in-sample optimization, as it takes into consideration partial trajectories that may not be relevant to the full trajectory. Formally:

\begin{align}
    &\ell_{RMS} (VO(A(\{I_t\}_{t=0}^L, P)), \{\delta_t^{t+1}\}_{t=0}^{L-1}) \nonumber \\
    &= \sum_{l=1}^L \ell_{VO} (VO(A(\{I_t\}_{t=0}^l, P)), \{\delta_t^{t+1}\}_{t=0}^{l-1})\\
    &\ell_{MPRMS} (VO(A(\{I_t\}_{t=0}^L, P)), \{\delta_t^{t+1}\}_{t=0}^{L-1}) \nonumber \\
    &= \sum_{l=1}^L \frac{1}{L-l+1} \sum_{i=0}^{L-l} \ell_{VO} (VO(A(\{I_t\}_{t=i}^{i+l}, P)), \{\delta_t^{t+1}\}_{t=i}^{i+l-1})
\end{align}

\section{Experiments}
\label{sec:exp}

\begin{figure*}
    \resizebox{\linewidth}{!}{
    \begin{tabular}{>{\centering\arraybackslash}m{2.5cm}|>{\centering\arraybackslash}m{4cm}>{\centering\arraybackslash}m{4cm}>{\centering\arraybackslash}m{4cm}>{\centering\arraybackslash}m{4cm}}  
    \textbf{Attack criteria} & \textbf{Opt RMS, Eval RMS} & \textbf{Opt RMS, Eval MPRMS} & \textbf{Opt MPRMS, Eval RMS} & \textbf{Opt MPRMS, Eval MPRMS} \\
    \hline
    \\
    \textbf{Synthetic data}& 
    \includegraphics[width=0.32\textwidth]{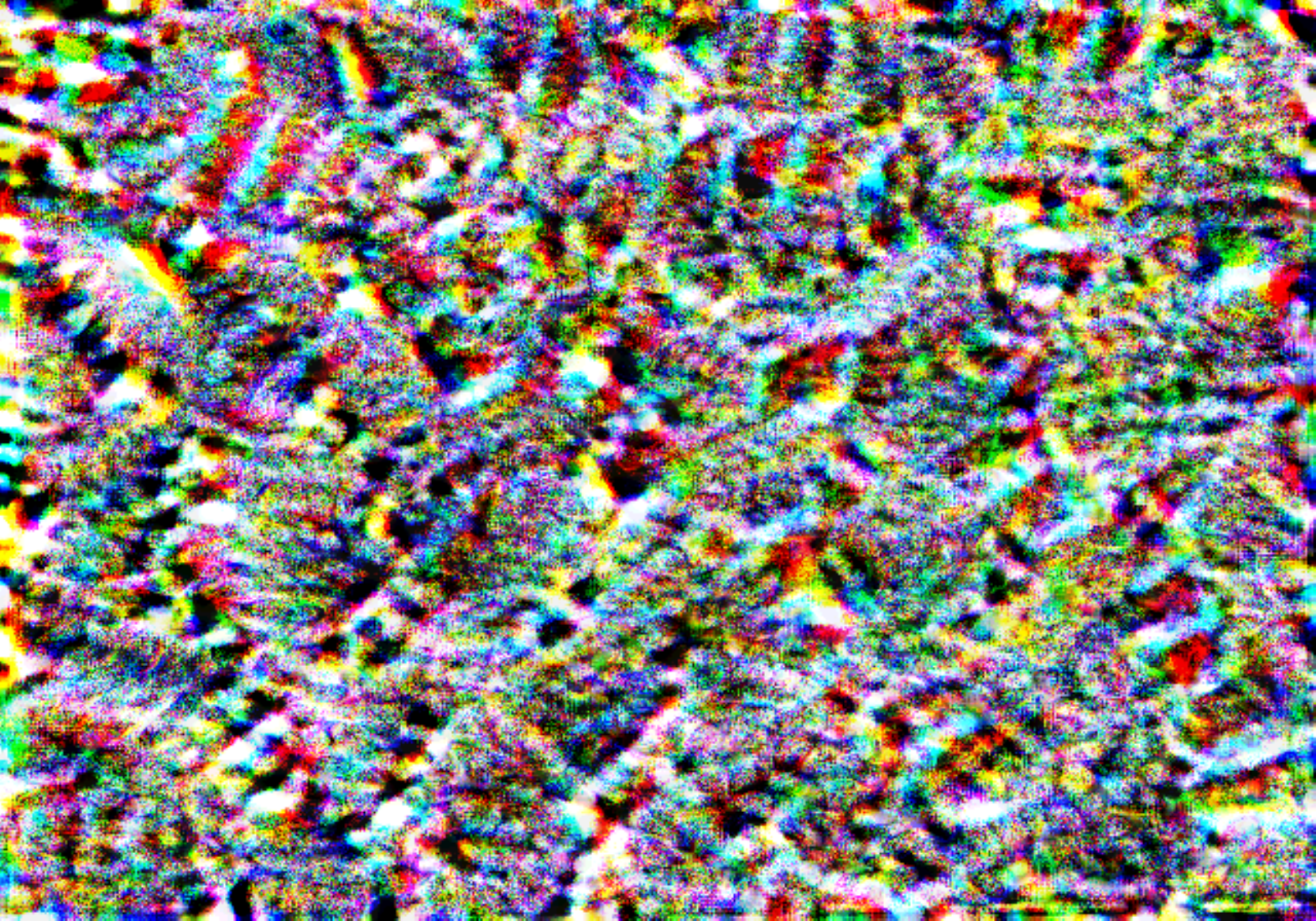}&
    \includegraphics[width=0.32\textwidth]{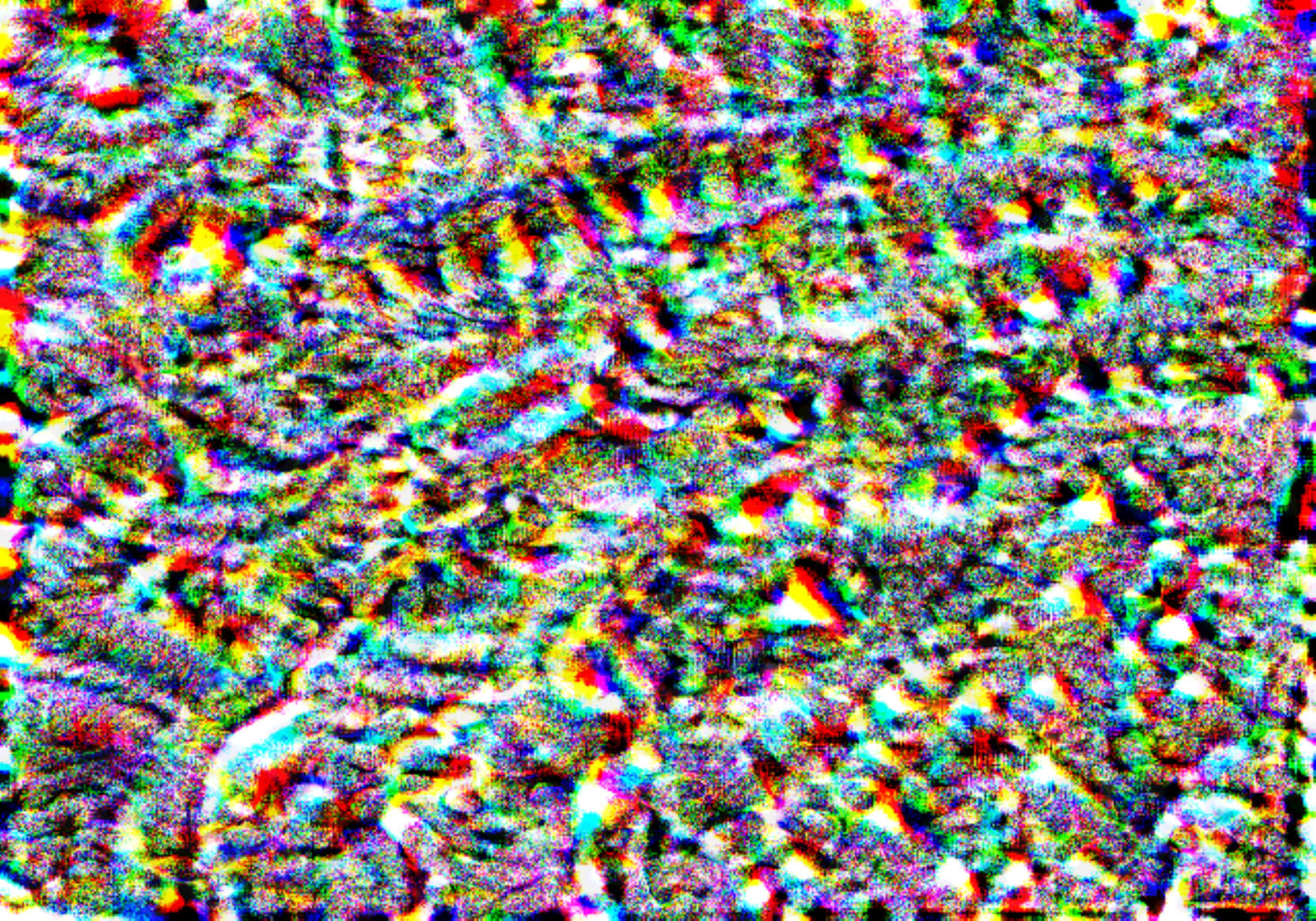}&
    \includegraphics[width=0.32\textwidth]{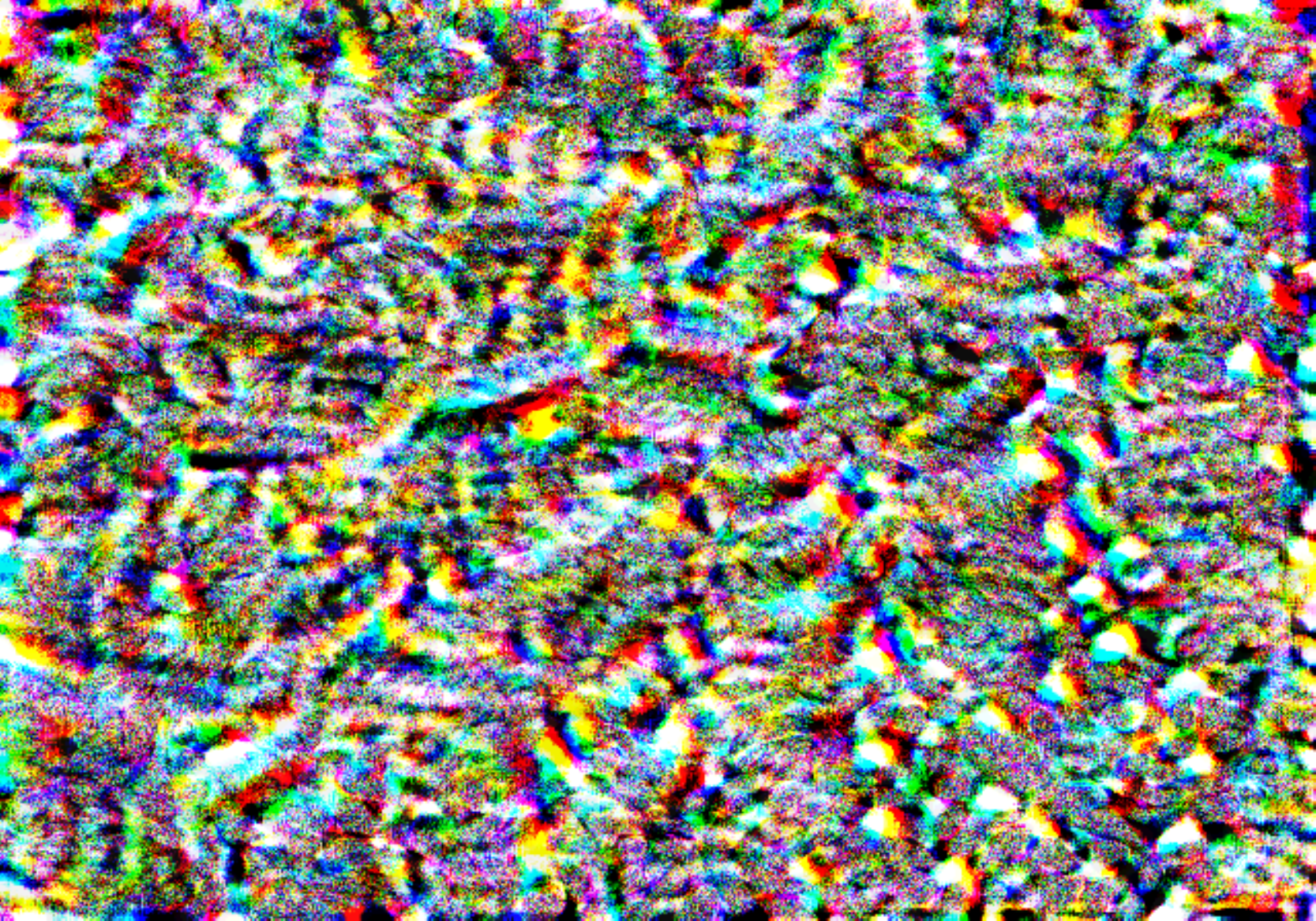}&
    \includegraphics[width=0.32\textwidth]{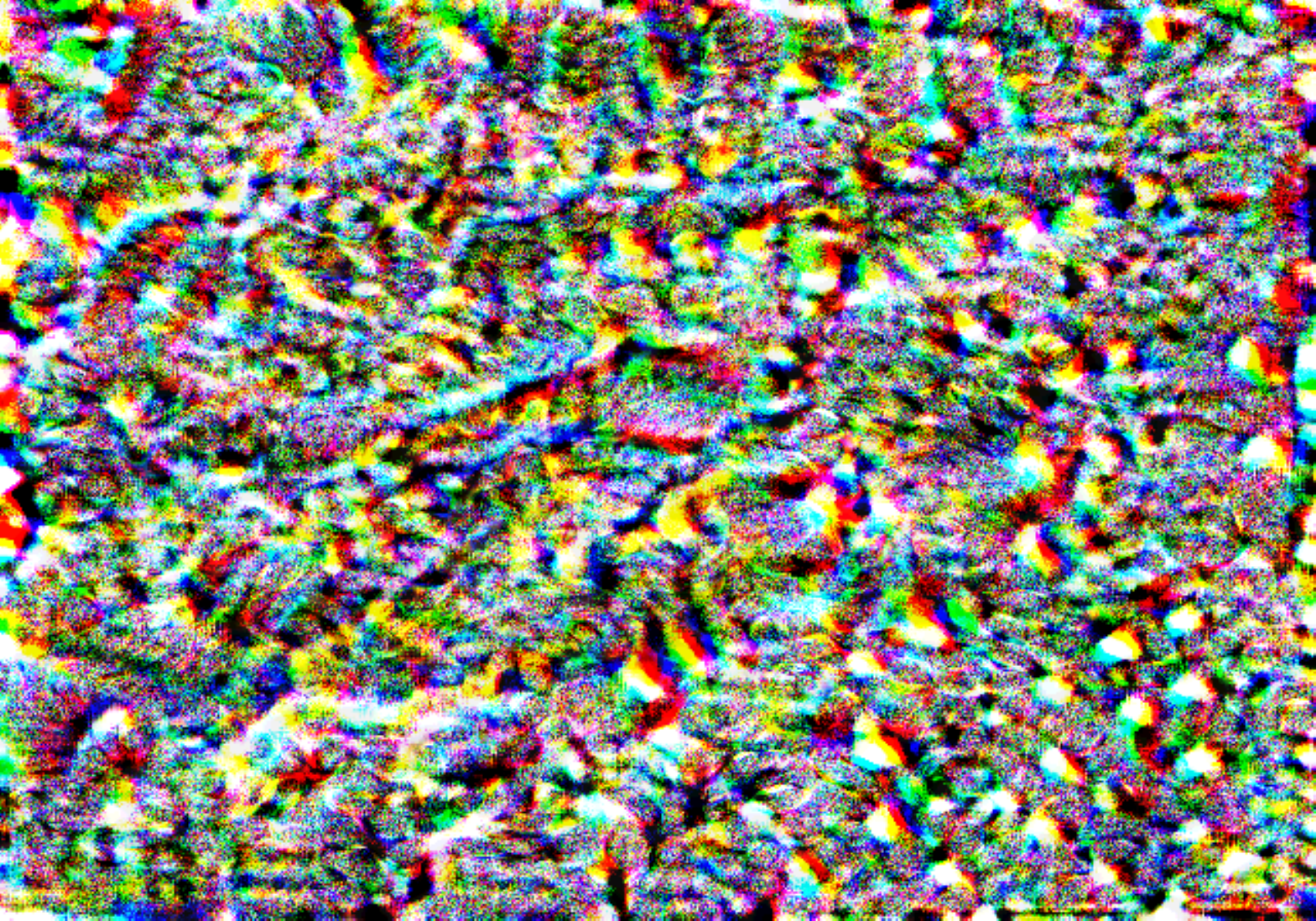} \\
     \textbf{Real data}&
    \includegraphics[width=0.32\textwidth]{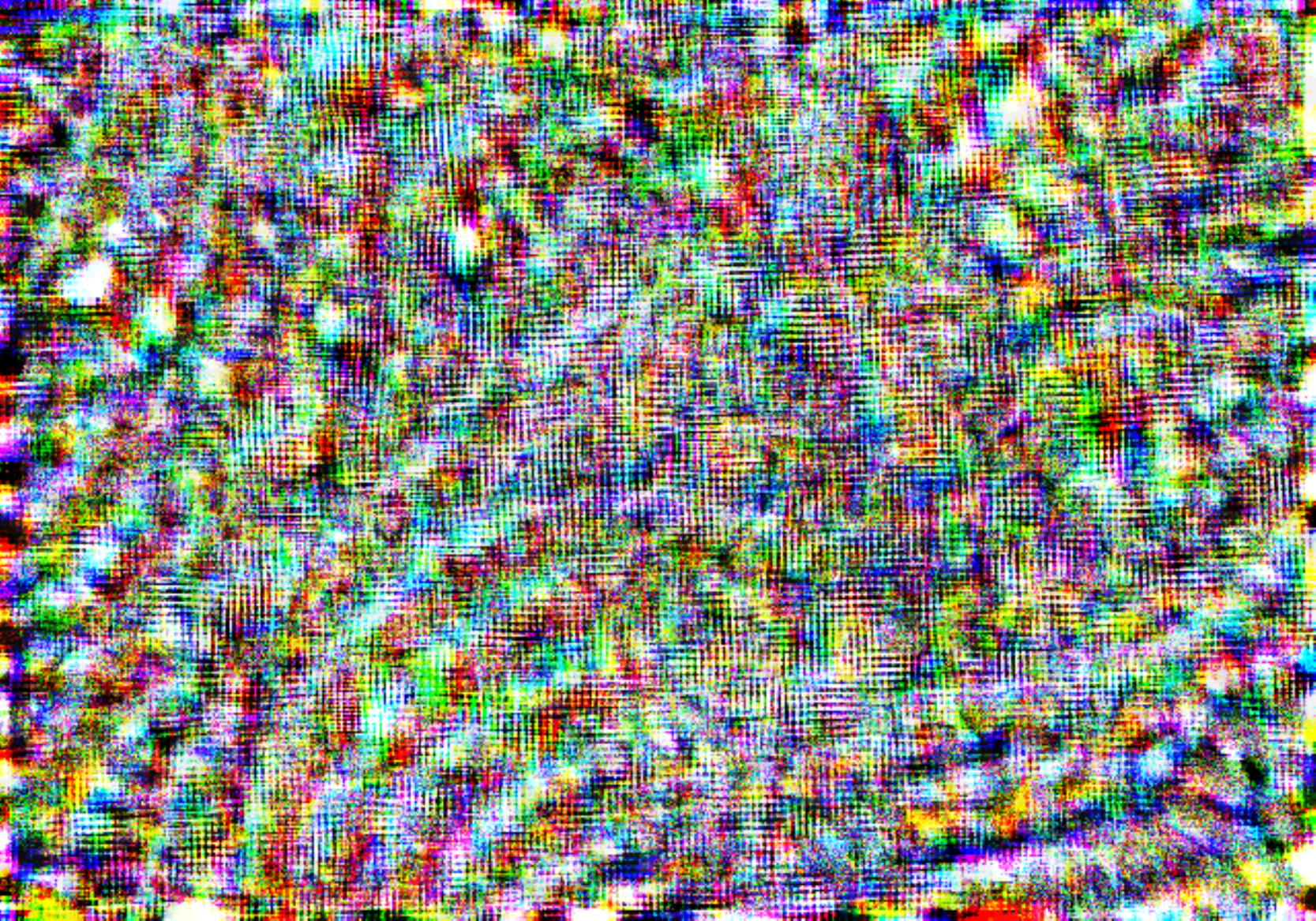}&
    \includegraphics[width=0.32\textwidth]{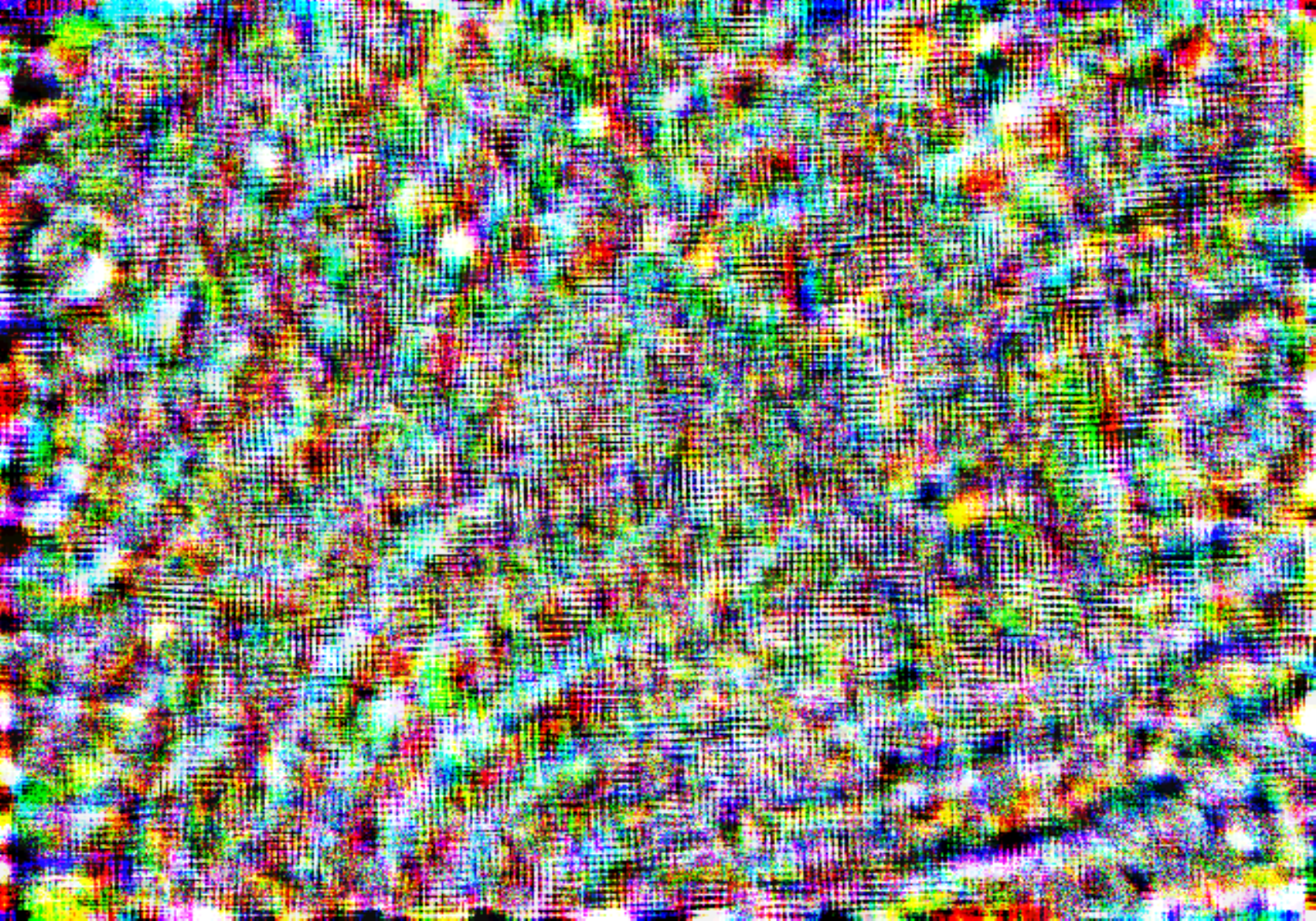}&
    \includegraphics[width=0.32\textwidth]{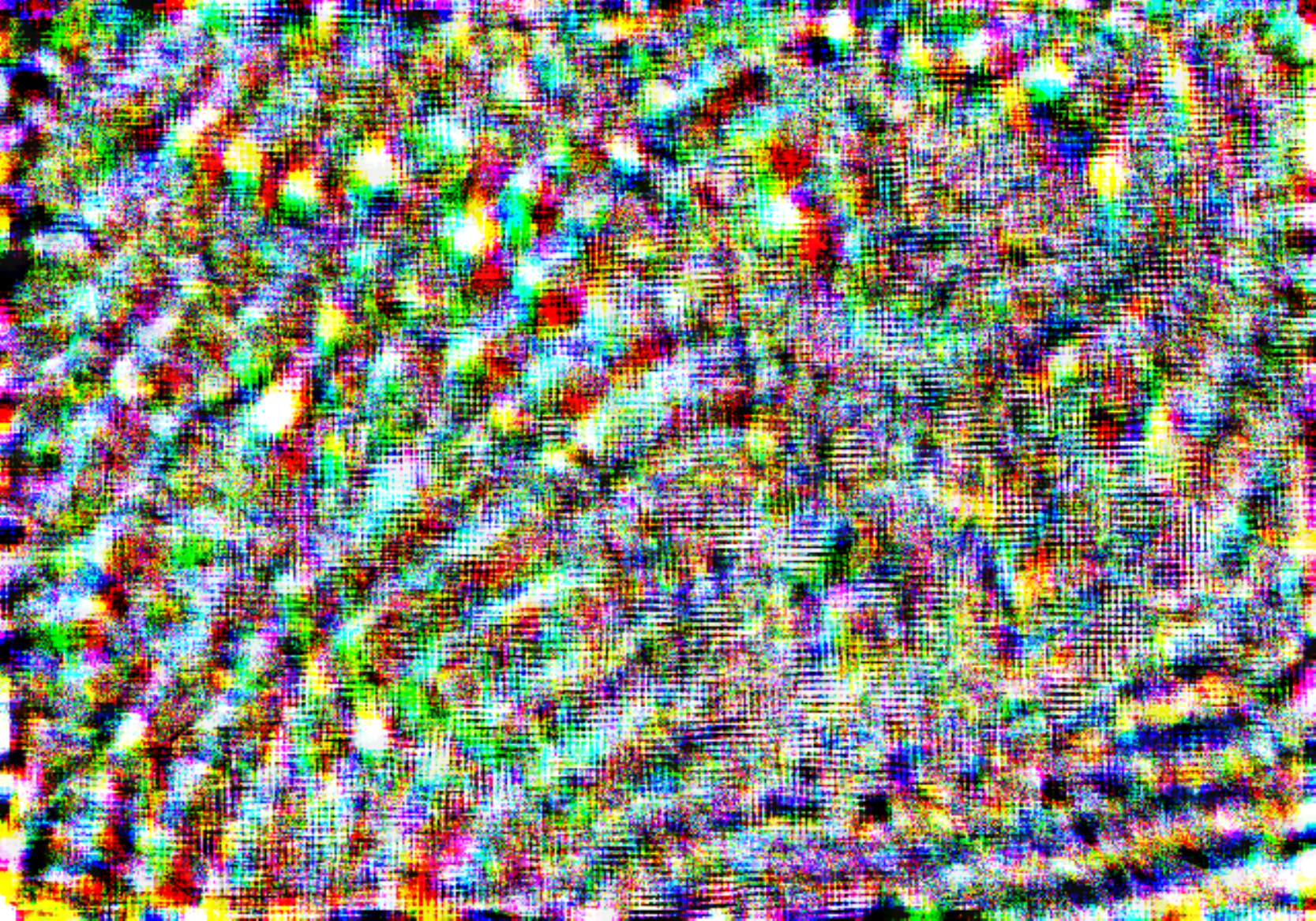}&
    \includegraphics[width=0.32\textwidth]{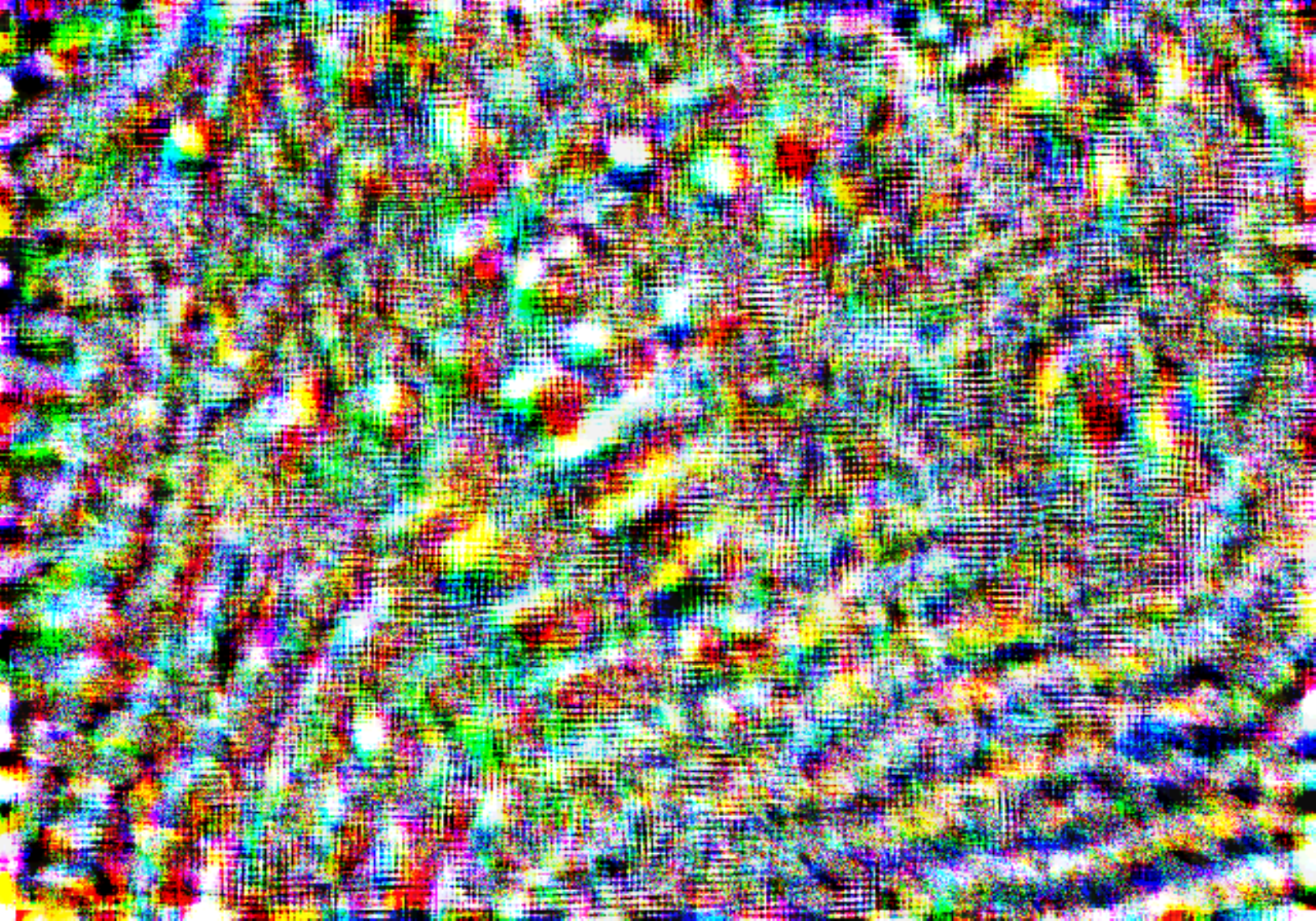} \\
    \end{tabular}
    }
    \caption{
    Visualization of universal adversarial patches. For each dataset, and optimization and evaluation criteria, we present the universal adversarial image produced via the in-sample attack scheme.}
    \label{fig:patch_vis}
\end{figure*}

We now present an empirical evaluation of the proposed method. We first describe the various experimental settings used for estimating the effect of the adversarial perturbations. We continue and describe the attacked VO model and the generation of both the synthetic and real datasets used in our experiments. Finally, we present our experimental results, first on the synthetic dataset and afterwards on the real dataset. 

\subsection{Experimental setting}
In this section we describe the experimental settings used for comparing the effectiveness of our method and various baselines. We differentiate between three distinct settings: in-sample, out-of-sample and closed-loop. For each experiment we report the average value of $\ell_{VO}$ between the estimated and ground truth motions over the test trajectories, compared to the length of the trajectory. We compare four methods of optimizing the attack to the clean results, by taking the training and evaluation criteria to be either $\ell_{RMS}$ or $\ell_{MPRMS}$. In all cases, we optimize the attacks for $k=100$ iterations.

\subsubsection{In-Sample}
The in-sample setting is used to estimate the effect of universal and PGD adversarial perturbations on known data. We train and test our attack on the entire dataset. We then compare the best performing attack to the random and clean baselines, for both our universal and PGD adversarial attacks. 

\subsubsection{Out-Of-Sample}
The out-of-sample setting is used to estimate generalization properties of universal perturbations to unseen data. Our methodology in this setting is to first split the trajectories into several folders, each with distinct initial positions of the contained trajectories. Thereafter, we perform cross-validation over the folders, where in each iteration a distinct folder is chosen to be the test set, and another to be the evaluation set. The training set thus comprises of the remaining folders. We report the average results over the test sets. Throughout our experiments, we use $10$-fold cross validation.

\subsubsection{Closed-Loop}
\label{sec:closed-loop}
The closed-loop setting is used to estimate the generalization properties of the previously produced adversarial patches to a closed-loop scheme, in which the outputs of the VO model are used in a simple navigation scheme. Our navigation scheme is an aerial path follower based on the carrot chasing algorithm \cite{perez2019aerial}. Given the current pose, target position and cruising speed, the algorithm computes a desired motion toward the target position. We then produce trajectories, each with a distinct initial and target position, with the motions computed iteratively by the navigation scheme based on the provided current position. The ground truth trajectories are computed by providing the current position in each step as the aggregation of motions computed by the navigation scheme. The estimated trajectories for a given patch, clean or adversarial, however, are computed by providing the current position in each step as the aggregation of motions estimated by the VO, where the viewpoint in each step corresponds to the aggregation of motions computed by the navigation scheme. We chose this navigation scheme to further assess the incremental effect of our adversarial attacks, as any deviation in the VO estimations directly affects the produced trajectory.

\subsection{VO model}
The VO model used in our experiments is the TartanVO \cite{wang2020tartanvo}, a recent differentiable VO model that achieved state-of-the-art performance in visual odometry benchmarks. Moreover, to better generalize to real-world scenarios, the model was trained over scale-normalized trajectories in diverse synthetic datasets. As the robustness of the model improves on scale-normalized trajectories, we supply it with the scale of the ground truth motions. The assumption of being aware of the motions' scale is a reasonable one, as the scale can be estimated to a reasonable degree in typical autonomous systems from the velocity. In our experiments, we found that the model yielded plausible trajectory estimates over the clean trajectories, for both synthetic and real data.

\subsection{Data generation}
\subsubsection{Synthetic data}
\begin{figure}
\begin{center}
\includegraphics[width=\linewidth]{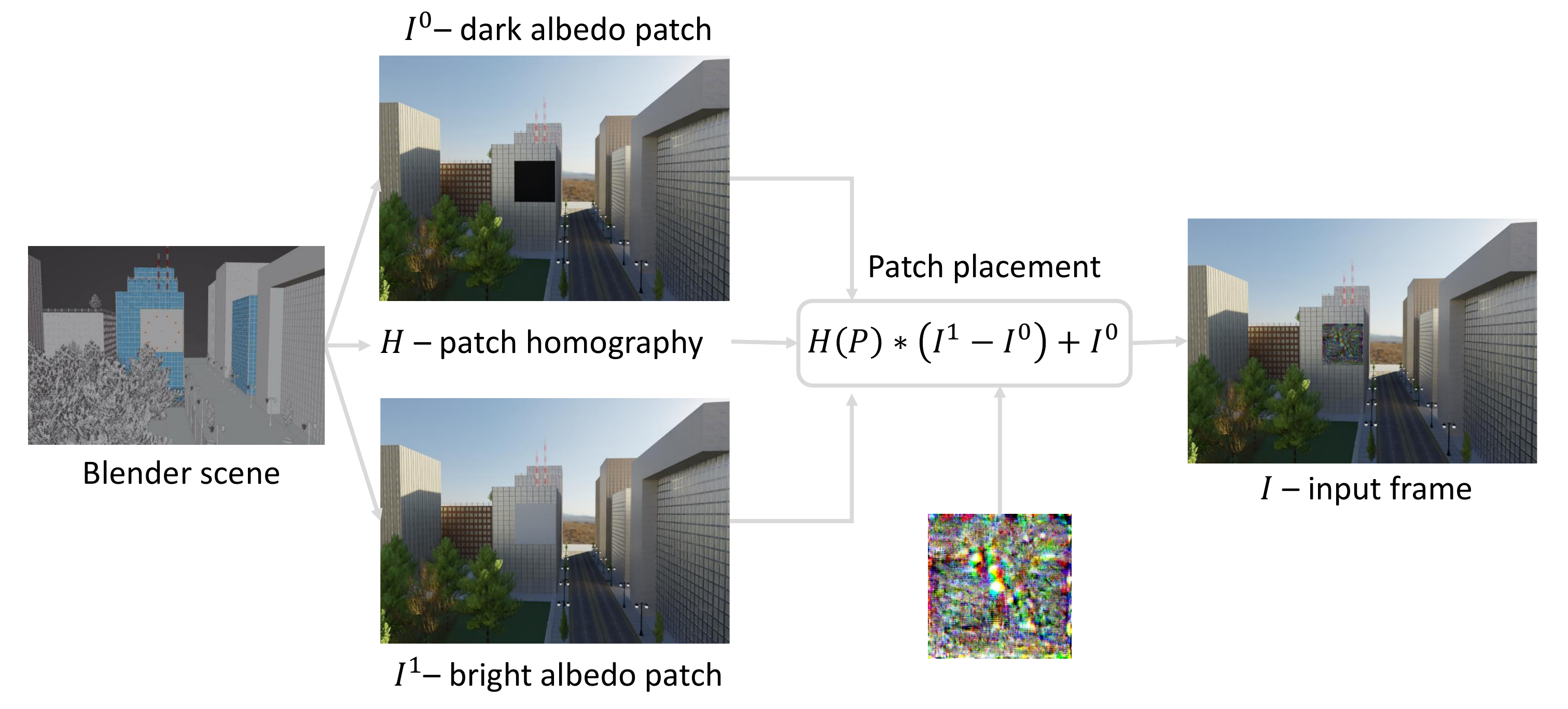}
\end{center}
       \caption{
       Synthetic frame generation. The attack patch $P$ is projected via the homography transformation $H$ and is incorporated into the scene according to the albedo images $I_0$ and $ I_1$.
       }
\label{fig:synth_frame}
\end{figure}

To accurately estimate the motions using the VO model, we require a photo-realistic renderer. In addition, the whole scene is altered for each camera motion, mandating re-rendering for each frame. An online renderer is impractical for optimization, in terms of computational overhead. An offline renderer is sufficient for our optimization schemes, and only our closed-loop test requires an online renderer. We, therefore, produce the data for optimizing the adversarial patches offline, and make use of an online renderer only for the closed-loop test. The renderer framework used is Blender \cite{blender}, a $3D$ modeling and rendering package. Blender enables photo-realistic rendered images to be produced from a given $3D$ scene along with the ground truth motions of the cameras. In addition, we produce high quality, occlusion-aware masks, which are then used to compute the homography transformation $H$ of the patch to the camera viewpoints. In the offline data generation of each trajectory $\{I_t\}_{t=0}^L$, we produce  $\{I_t^0\}_{t=0}^L, \{I_t^1\}_{t=0}^L, \{H_t\}_{t=0}^L$ , as in \cref{eq:adv_pert}, as well as the ground truth camera motions $\delta_t^{t+1}$. For the closed-loop test, for each initial position, target position and pre-computed patch $P$, we compute the ground truth motions $\delta_t^{t+1}$ and their estimation by the VO model $\hat{\delta}_t^{t+1}$, as described in \cref{sec:closed-loop}. The frame generation process is depicted in \cref{fig:synth_frame}. We produced the trajectories in an urban $3D$ scene, as in such a scenario, GPS reception and accuracy is poor, and autonomous systems rely more heavily on visual odometry for navigation purposes. The patch is then positioned on a square plane at the side of one of the buildings, in a manner that resembles a large advertising board.

\paragraph{Offline rendered data specifics}
We produced $100$ trajectories with a constant linear velocity norm of $v=5[\frac{m}{s}]$, and a constant $2D$ angular velocity sampled from $v_{\theta}=\mathcal{N}(0, 3) [\frac{{deg}}{s}]$. Each trajectory is nearly $10[m]$ long and contains $60$ frames at $30$ fps. The trajectories are evenly divided between $10$ initial positions, with the initial positions being distributed evenly on the ring of a right circular cone with a semi-vertical angle of $10^\circ$ and a $50[m]$-long axis aligned with the patch's normal. We used a camera with a horizontal field-of-view (FOV) of $80^{\circ}$ and $640\times448$ resolution. The patch is a $30[m]$ square, occupying, under the above conditions, an average FOV over the trajectories ranging from $18.1\%$ to $27.3\%$, and covering a mean $22.2\%$ of the images. To estimate the effect of the patch's size, which translates into a $\ell_0$ limitation on the adversarial attacks, we also make use of smaller sized patches. The outer margins of the patch would then be defaulted to the clean $I_0$ image, and the adversarial image would be projected onto a smaller sized square, with its center aligned as before.

\paragraph{Closed-loop data specifics}

Similarly, in the closed-loop scheme we produced trajectories with the same camera and patch configuration, the same distribution of initial positions and with the navigation scheme cruising speed set according to the previous linear velocity  norm of $v=5[\frac{m}{s}]$. Here we, however, produce $10$ trajectories by randomly selecting a target position at the proximity of the patch for each initial position. We then produced the ground truth and VO trajectories for each patch $P$ as described in \cref{sec:closed-loop}. The trajectories are each $45[m]$ long and contain $270$ frames at $30$ fps.

\subsubsection{Real data}
\begin{figure*}
    \centering
    \resizebox{\linewidth}{!}
    {
    \begin{tabular}{cccc}
    \includegraphics[width=0.25\textwidth]{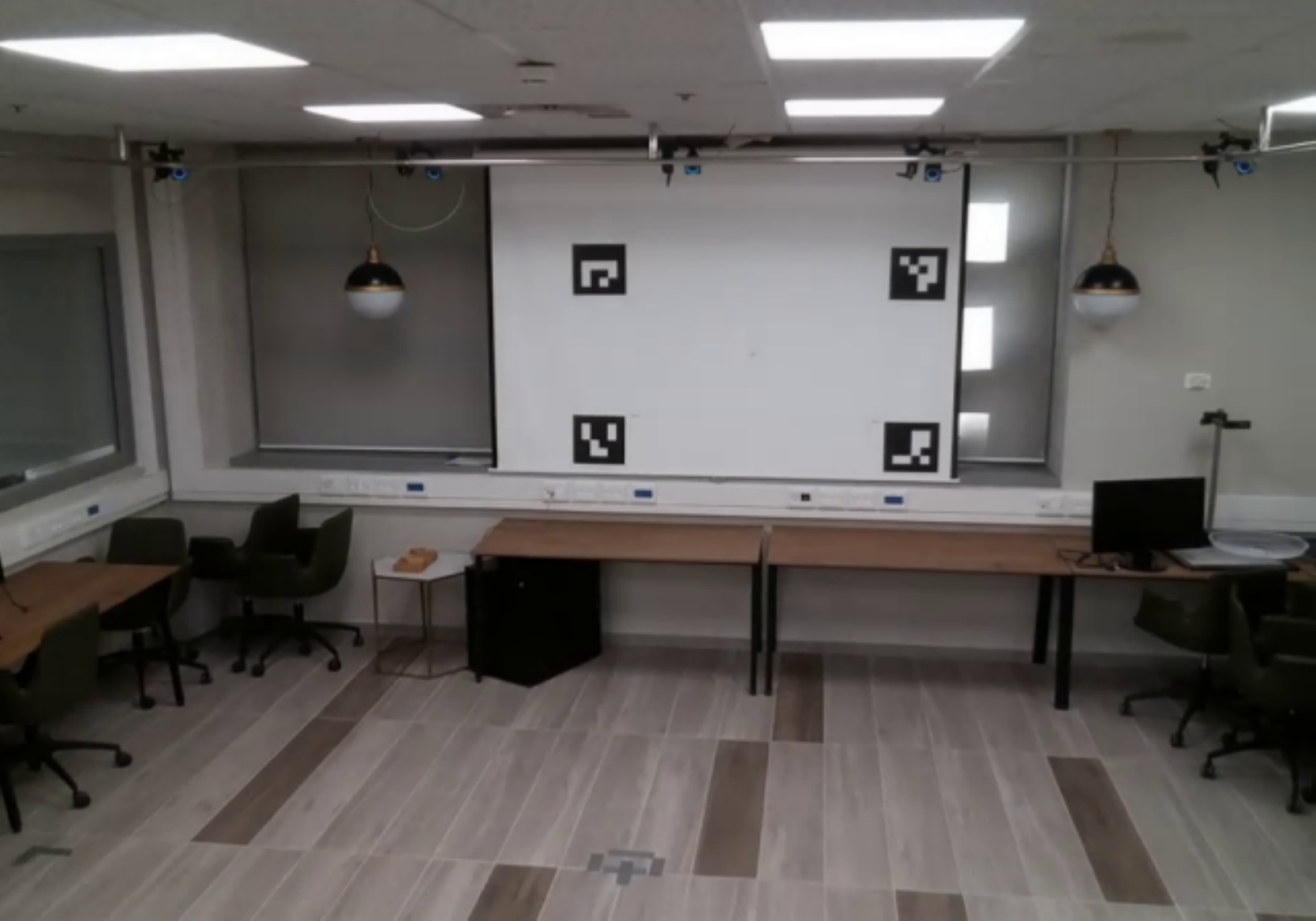}&
    \includegraphics[width=0.25\textwidth]{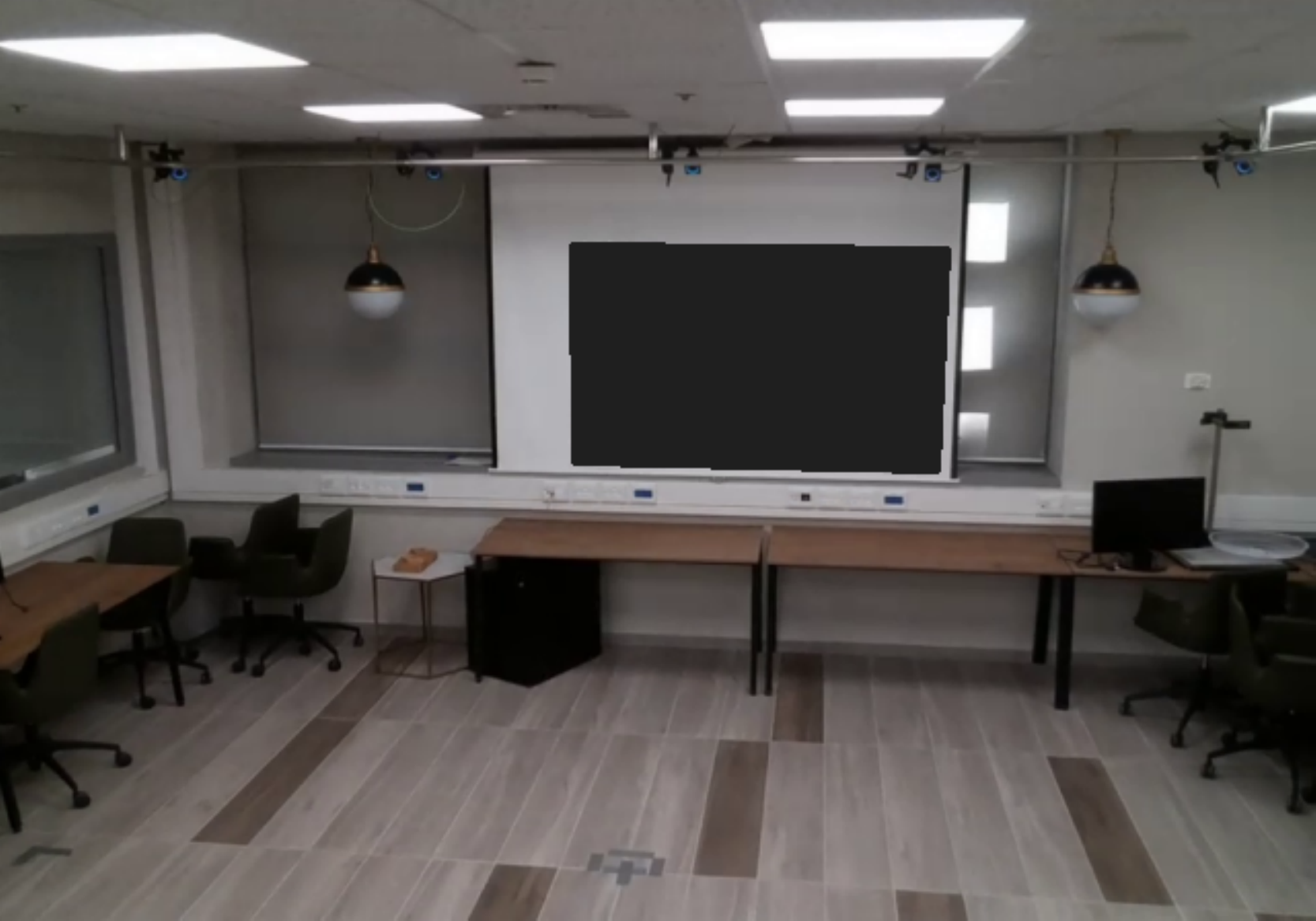}&
    \includegraphics[width=0.25\textwidth]{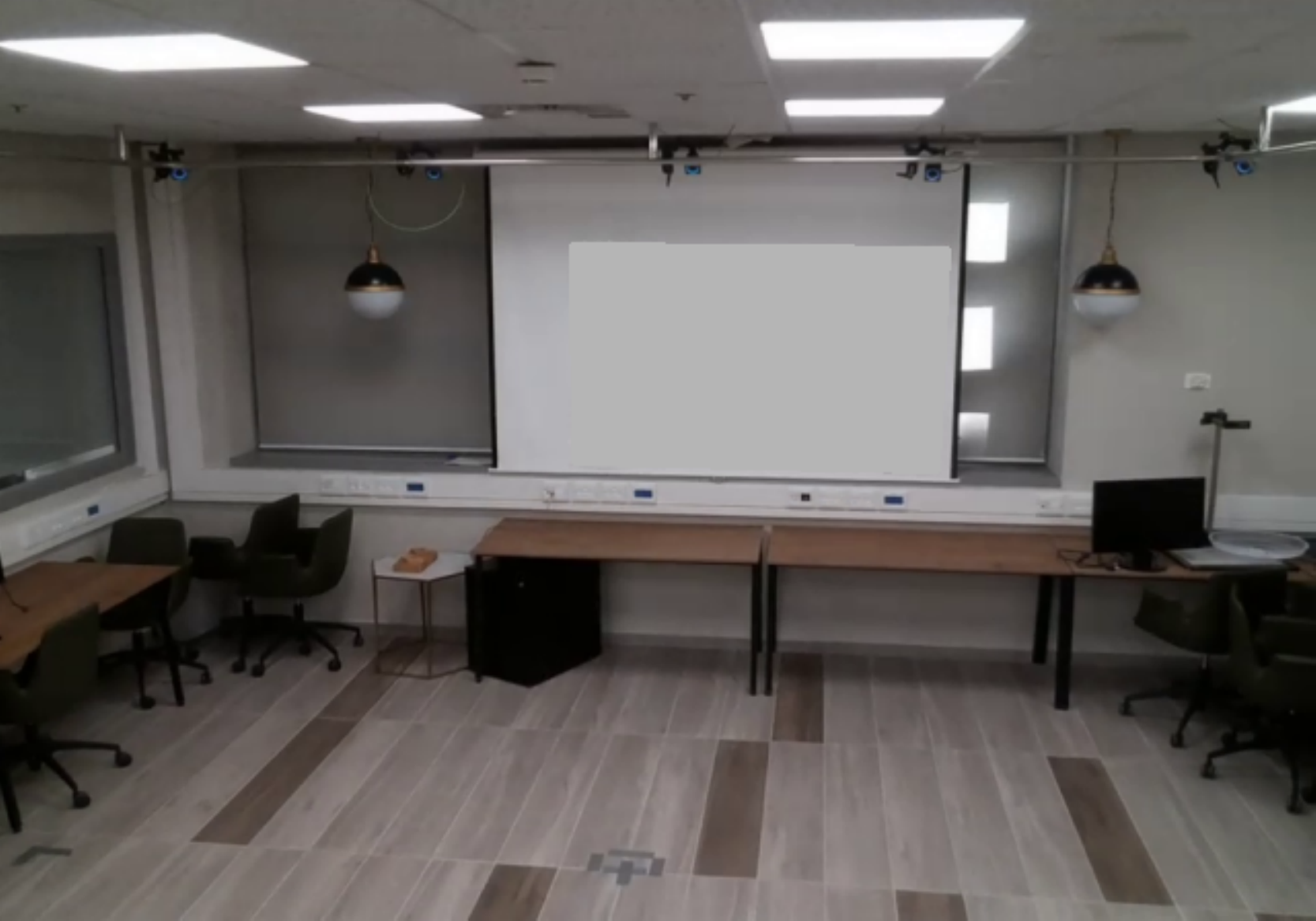}&
    \includegraphics[width=0.25\textwidth]{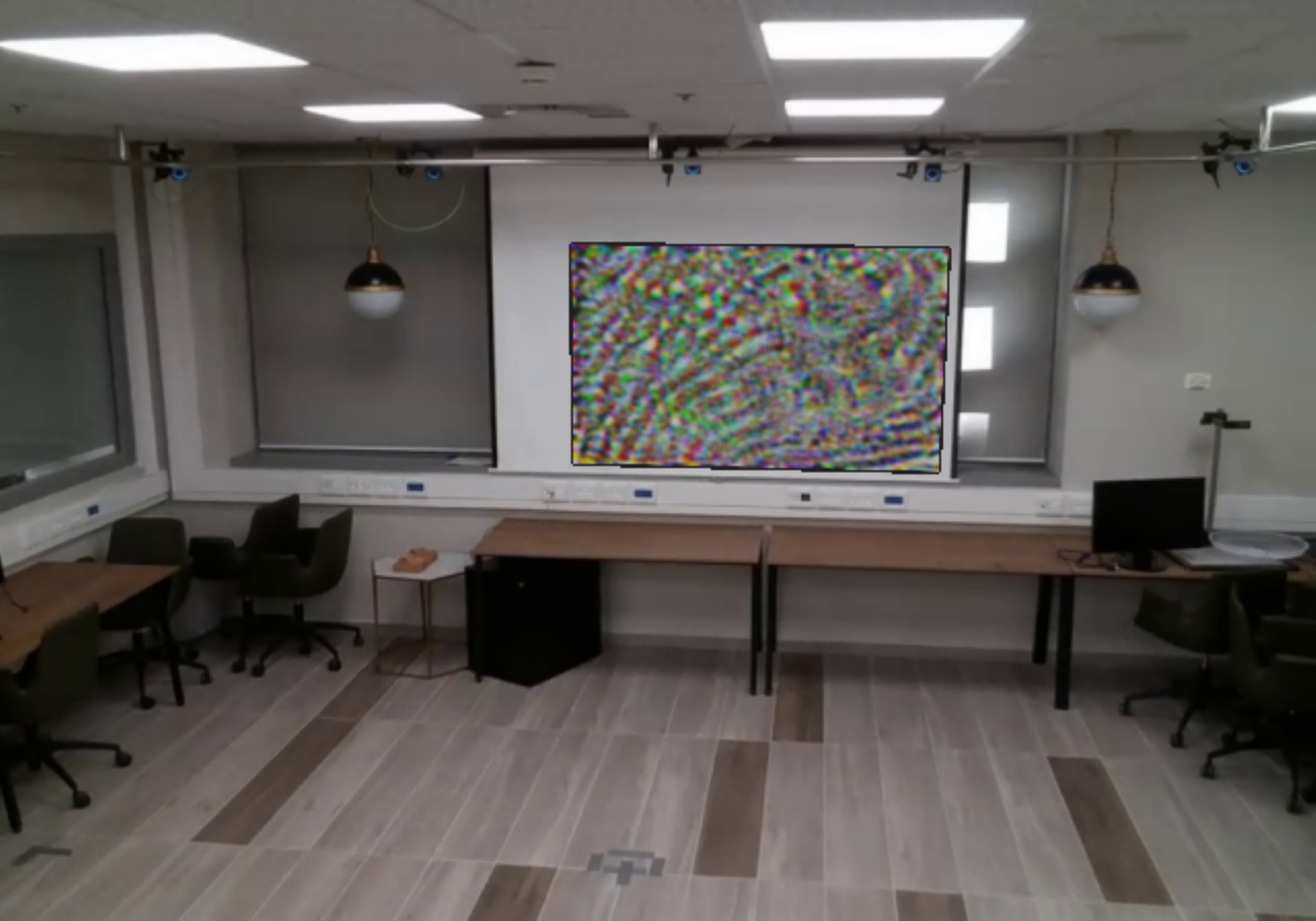}
    \\
    (a) &
    (b) &
    (c) &
    (d)
    \end{tabular}
    }
    \caption{
    Real dataset frame generation. (a) Original image. (b+c) Black and white albedo approximations. (d) Adversarial patch projected onto the scene.
    }
    \label{fig:real_frame}
\end{figure*}

In the real data scenario, we situated a DJI Tello drone inside an indoor arena, surrounded by an Optitrack motion capture system for recording the ground truth motions. The patch was positioned on a planar screen at the arena boundary. To compute the homography transformation $H$ of the patch to the camera viewpoints, we designated the patch location in the scene by four Aruco markers. Similarly to the offline synthetic data generation, for each trajectory $\{I_t\}_{t=0}^L$ we produced $\{I_t^0\}_{t=0}^L, \{I_t^1\}_{t=0}^L, \{H_t\}_{t=0}^L$ as well as the ground truth camera motions $\delta_t^{t+1}$.
An example data frame is depicted in \cref{fig:real_frame}.

We produced $48$ trajectories with a constant velocity norm of approximately $v \simeq 1[\frac{m}{s}]$. Each trajectory contained $45$ frames at $30$ fps with total length $ l\sim \mathcal{N}(1.56, 0.15^2) [m]$. The trajectories' initial positions were distributed evenly on a plane parallel to the patch at a distance of $7.2[m]$. Not including the drone movement model, the trajectories comprised linear translation toward evenly distributed target positions at the patch's plane. We used a camera with a horizontal FOV of $82.6^{\circ}$, and $640\times448$ resolution. The patch was a $1.92\times1.24[m]$ rectangle, occupying, under the above conditions, an average FOV over the trajectories ranging from $6.8\%$ to $11.2\%$, and covering a mean $8.8\%$ of the images.

\subsection{Experimental results}
\subsubsection{Synthetic data experiments}
\begin{figure*}
 \centering
    \includegraphics[width=0.483\linewidth]{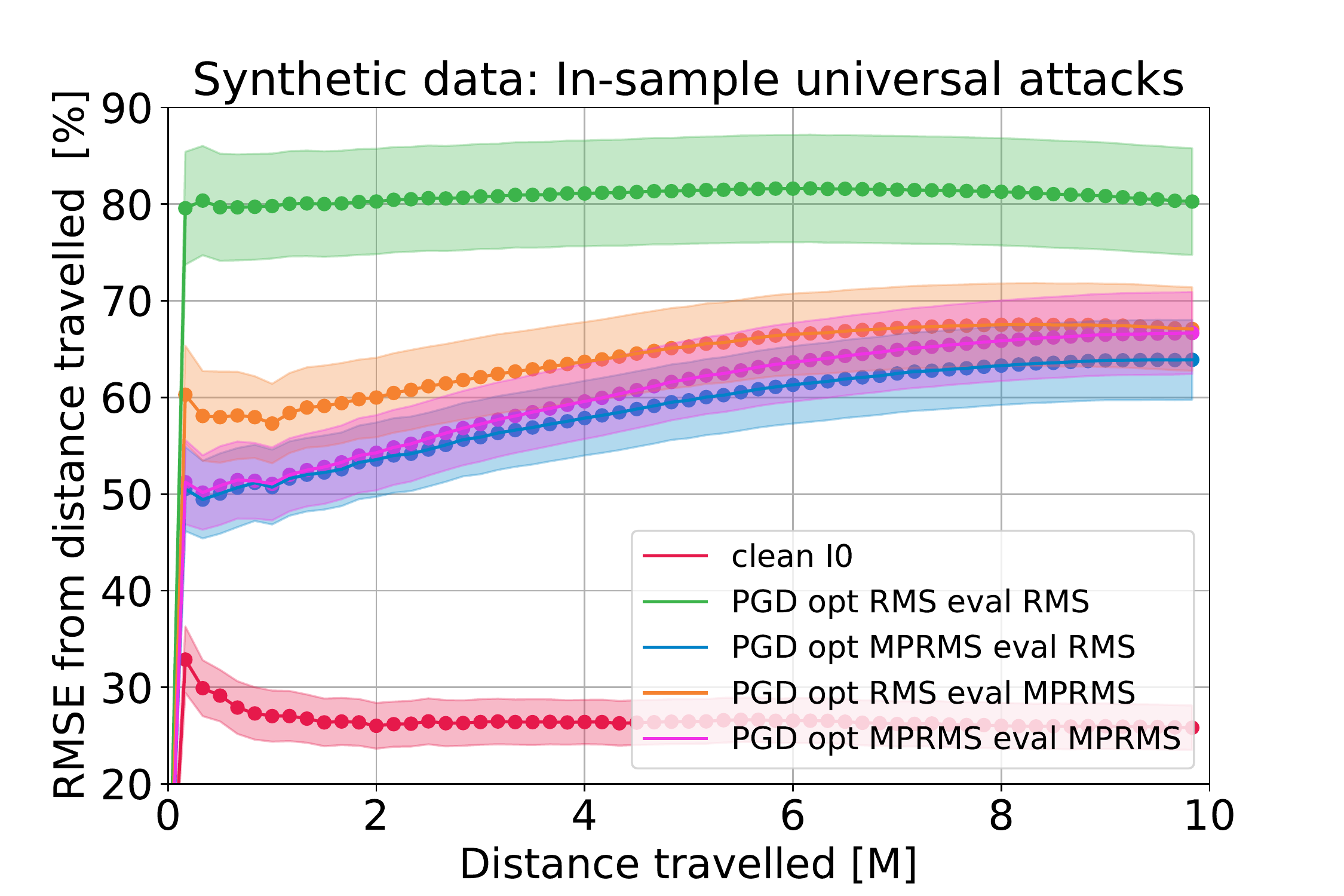}
    \includegraphics[width=0.483\linewidth]{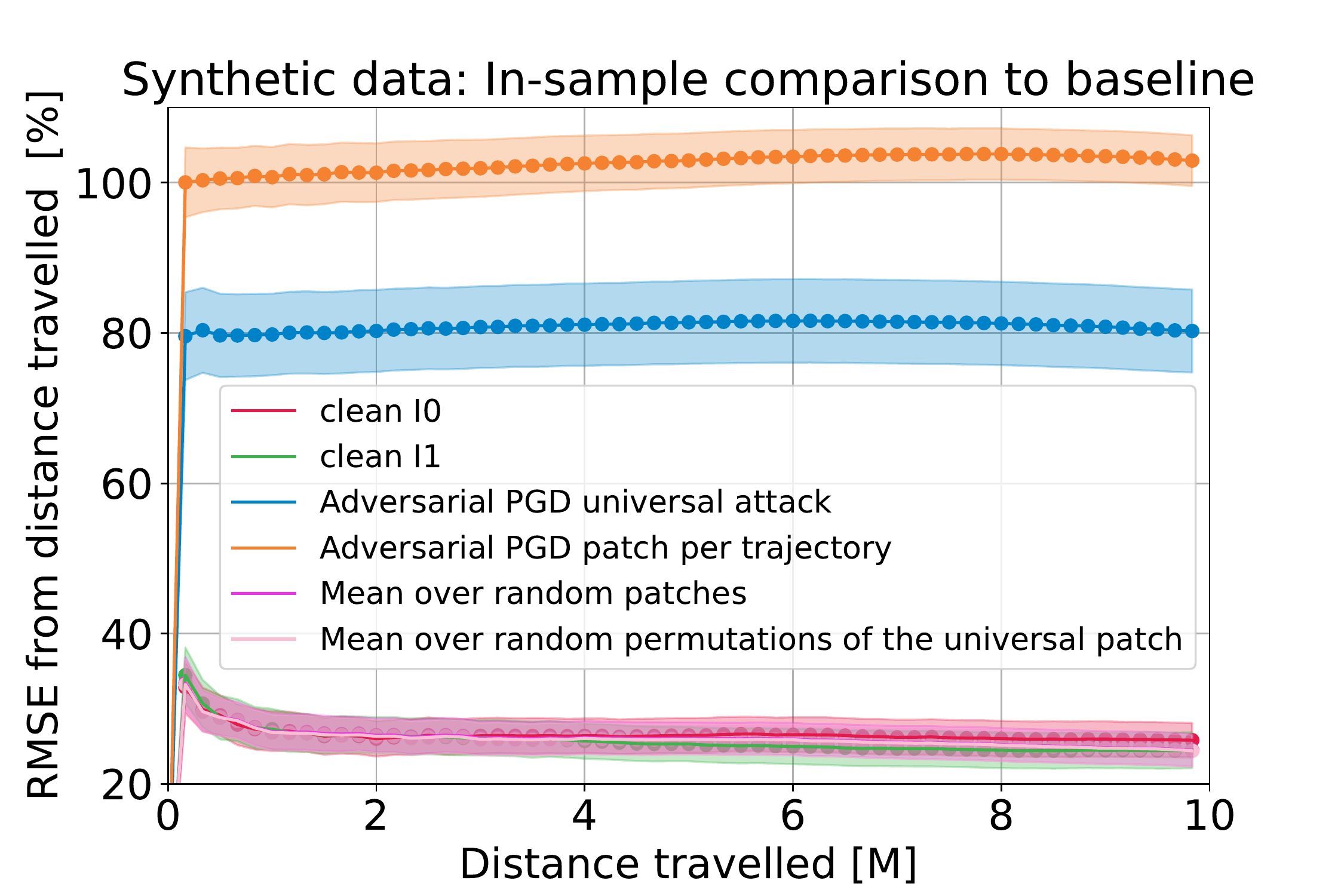}
 \caption{
 Accumulated deviation in distance travelled from the ground-truth trajectories of the synthetic dataset as a function of the trajectory length. We show a comparison of our universal attacks trained on the entire dataset (left), and a comparison of our best performing universal and PGD attacks to the clean and random perturbation baselines (right). We present mean and standard deviation over the trajectories for each trajectory length.
 }
\label{fig:synth_in_sample}
\end{figure*}

In \cref{fig:synth_in_sample} we show the in-sample results on the synthetic dataset. Both our universal and PGD attacks showed a substantial increase in the generated deviation over the clean and random baselines. The best PGD attack generated, after $10[m]$, a deviation of $103\%$ in distance travelled, which is a factor of $399\%$ from the clean $I^0$ baseline. For the same configuration, the best universal attack generated a deviation of $80\%$ in distance travelled, which is a factor of $311\%$ from the clean $I^0$ baseline. Moreover, the clean $I^1$ and random baselines show a slight decrease in the generated deviation over the clean $I^0$ results, including the random permutations of the best universal patch. This suggests that the VO model is affected by the structure of the adversarial patch rather than simply by the color scheme. 
In addition, for both the universal and PGD attacks, the best performance was achieved for $\ell_{train}=\ell_{RMS}$, where the PGD attacks showed negligible change in the choice of evaluation criterion, and $\ell_{eval}=\ell_{RMS}$ is clearly preferred for universal attacks. This supports our assumption that $\ell_{MPRMS}$ may be less suited for in-sample optimization.\\
\begin{figure*}
 \centering
    \includegraphics[width=0.483\linewidth]{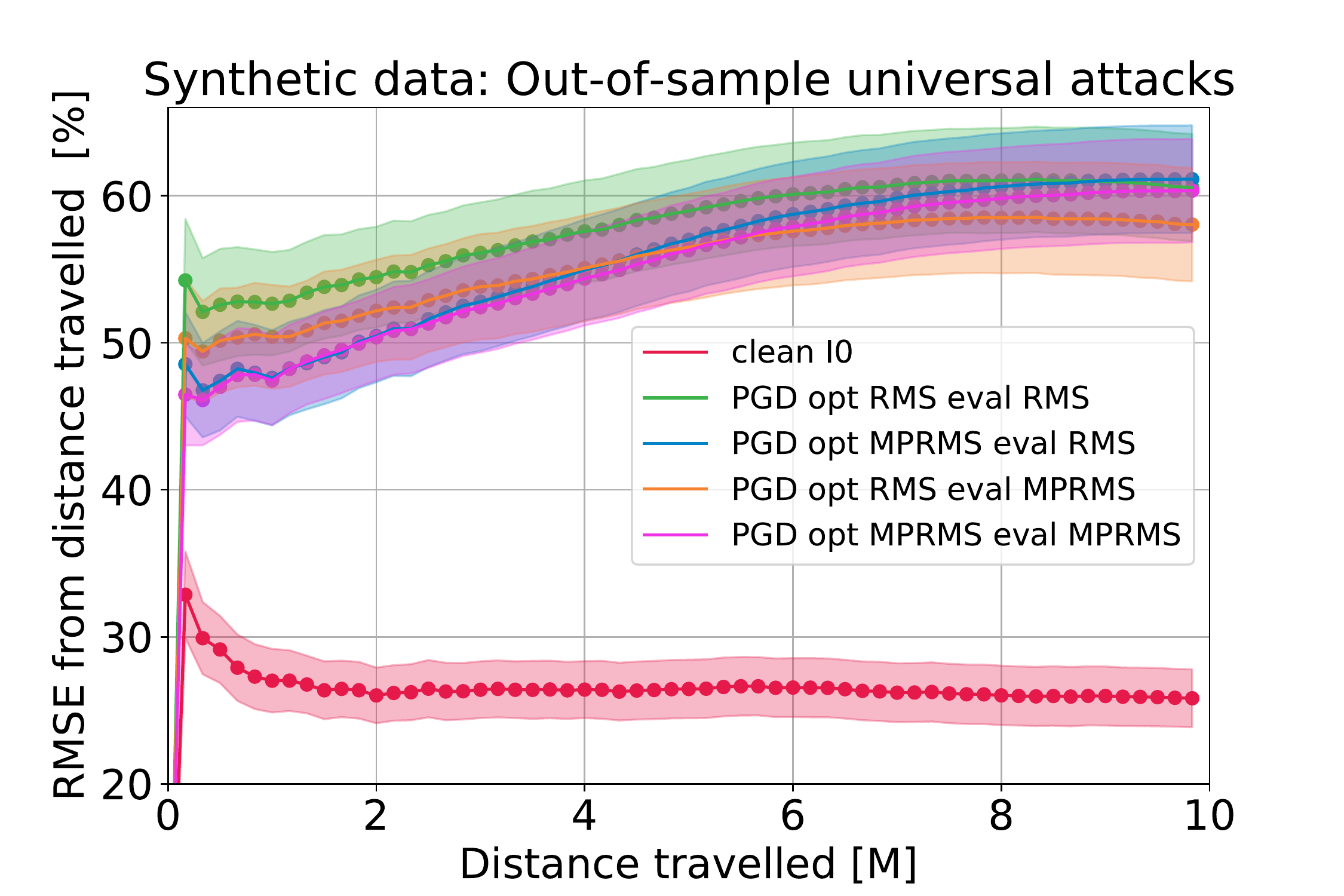}
   \includegraphics[width=0.483\linewidth]{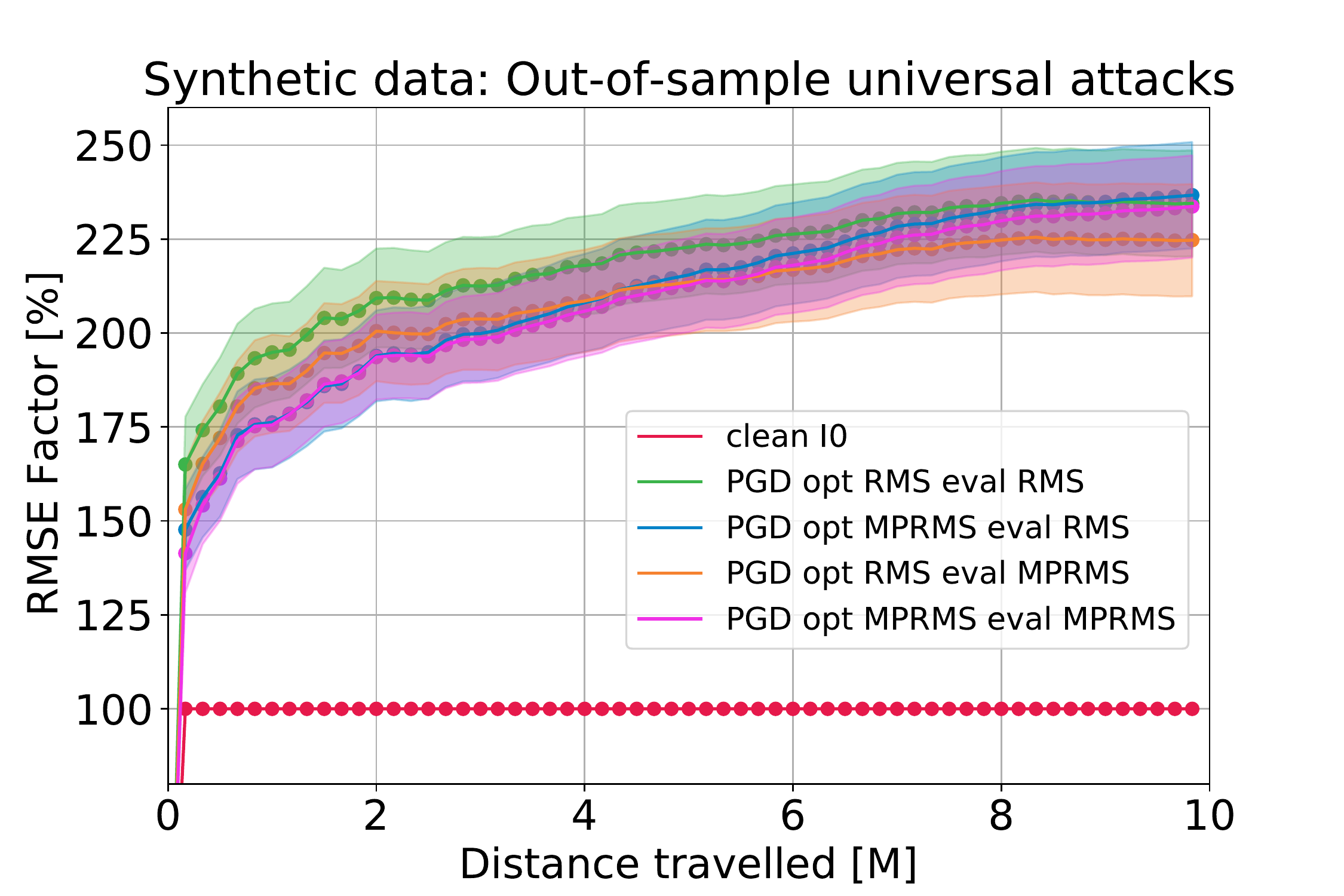}
 \caption{
  Accumulated deviation in distance travelled from the ground-truth trajectories over out-of-sample cross-validation of the synthetic dataset as a function of the trajectory length. We show a comparison of the deviation in distance travelled between our universal attacks and the clean baseline (left) as well as the ratio of the deviation compared to the clean results (right). We present mean and standard deviation over the trajectories for each trajectory length.
 }
\label{fig:synth_oos}
\end{figure*}

In \cref{fig:synth_oos} we show the out-of-sample results on the synthetic dataset. Our universal attacks again showed a substantial increase in the generated deviation over the clean baseline, with the best universal attack generating, after $10[m]$, a deviation of $61\%$ in distance travelled, which is a factor of $237\%$ from the clean $I^0$ baseline. As for the choice of criteria, the best performance is achieved for the $\ell_{train}=\ell_{MPRMS}$ optimization criterion with a slight improvement of $\ell_{eval}=\ell_{RMS}$ as the evaluation criterion. This supports our assumption that $\ell_{MPRMS}$ is better suited for generalization to unseen data, and may indicate that $\ell_{RMS}$ is better suited for evaluation.\\
\begin{figure*}
 \centering
    \includegraphics[width=0.483\linewidth]{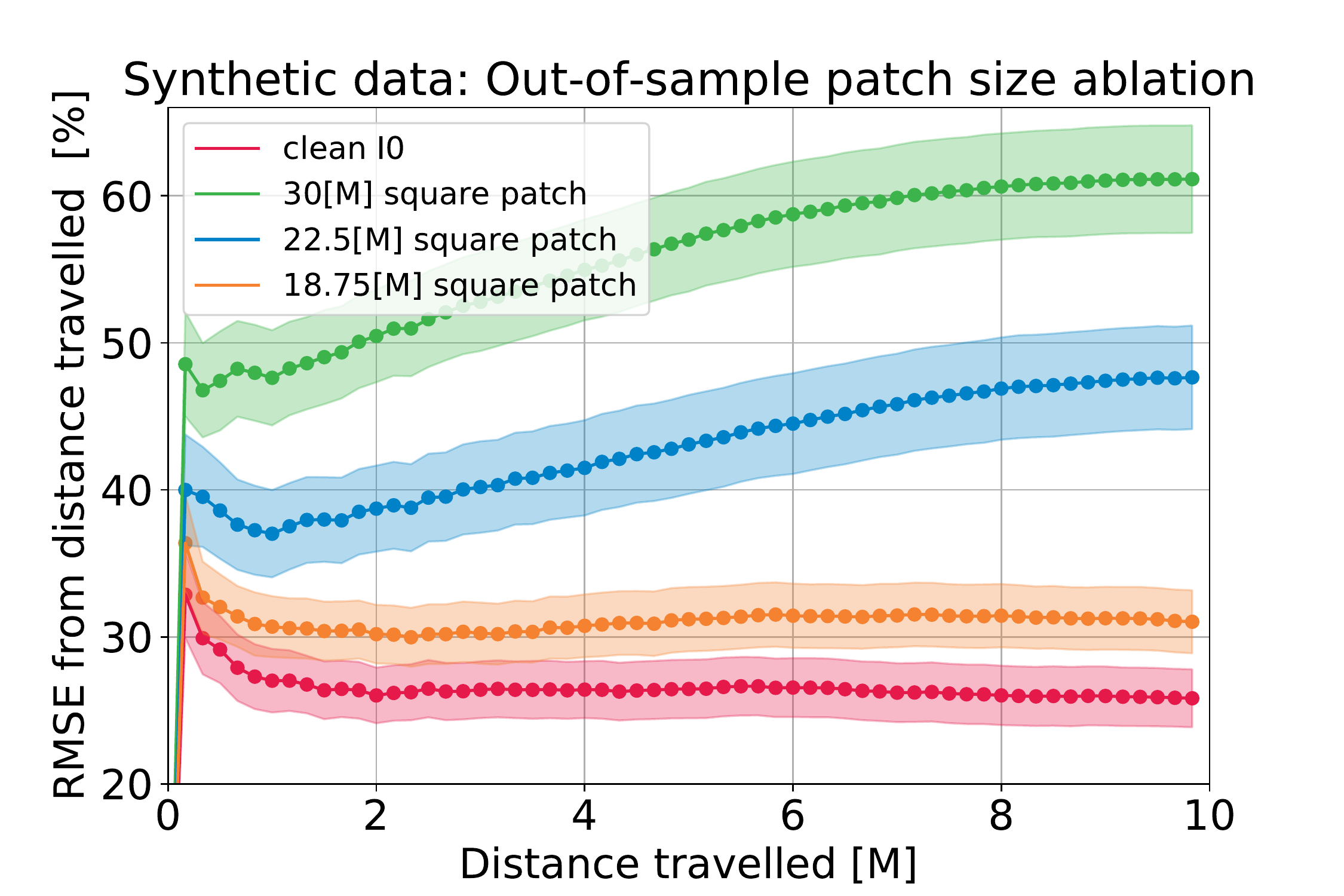}
    \includegraphics[width=0.483\linewidth]{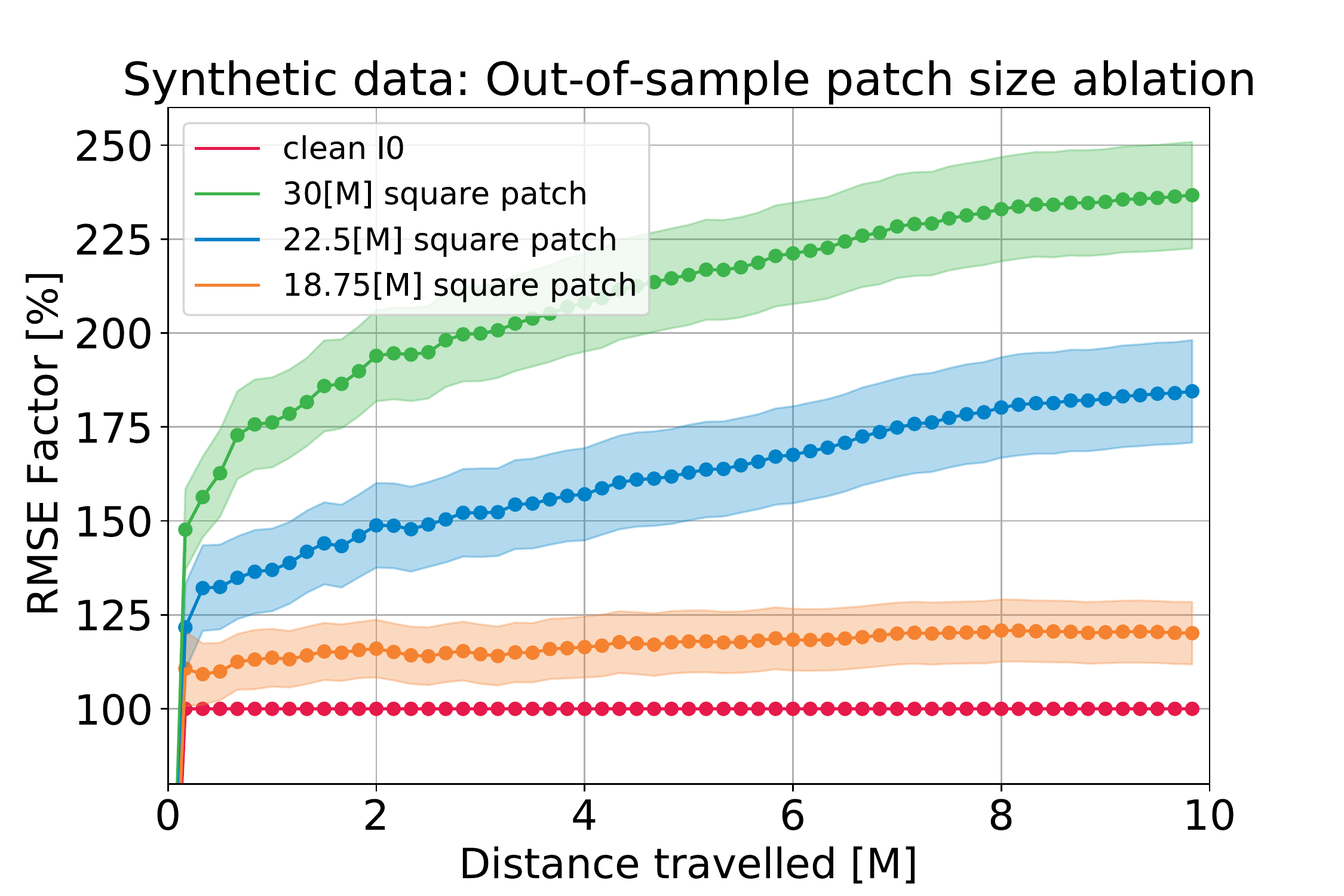}
 \caption{
   A comparison of different patch sizes of the accumulated deviation in distance travelled from the ground-truth trajectories over out-of-sample cross-validation of the synthetic dataset as a function of the trajectory length. The patches are a $30[m]$, a $22.5[m]$, and $18.75[m]$ squares and occupy a FOV over the trajectories ranging from $18.1\%-27.4\%, 8.3\%-12.6\%, 6.1\%-9.3\%$ respectively, and covering a mean $22.2\%, 10.2\%, 7.5\%$ of the images. We show a comparison of the deviation in distance travelled between our best performing universal attacks for each patch and the clean baseline (left) as well as the ratio of the deviation compared to the clean results (right). We present mean and standard deviation over the trajectories for each trajectory length.
 }
\label{fig:synth_fov}
\end{figure*}

In \cref{fig:synth_fov} we show the patch size comparison of the out-of-sample results on the synthetic dataset. The best performance for all patch sizes is achieved for the same choice of optimization and evaluation criteria, which supports our previous indications. Nevertheless, as the patch size is reduced, the increase in the generated deviation becomes less significant. For the $22.5[m]$ square patch, the best performing universal attack generated, after $10[m]$, a deviation of $48\%$ in distance travelled, which is a factor of $184\%$ from the clean baseline. Regarding the $18.75[m]$ square patch, the generated deviation decays significantly with the best performing universal attack generating, after $10[m]$, a deviation of $31\%$ in distance travelled, which is a factor of $120\%$ from the clean baseline.\\
\begin{figure*}
 \centering
    \includegraphics[width=0.483\linewidth]{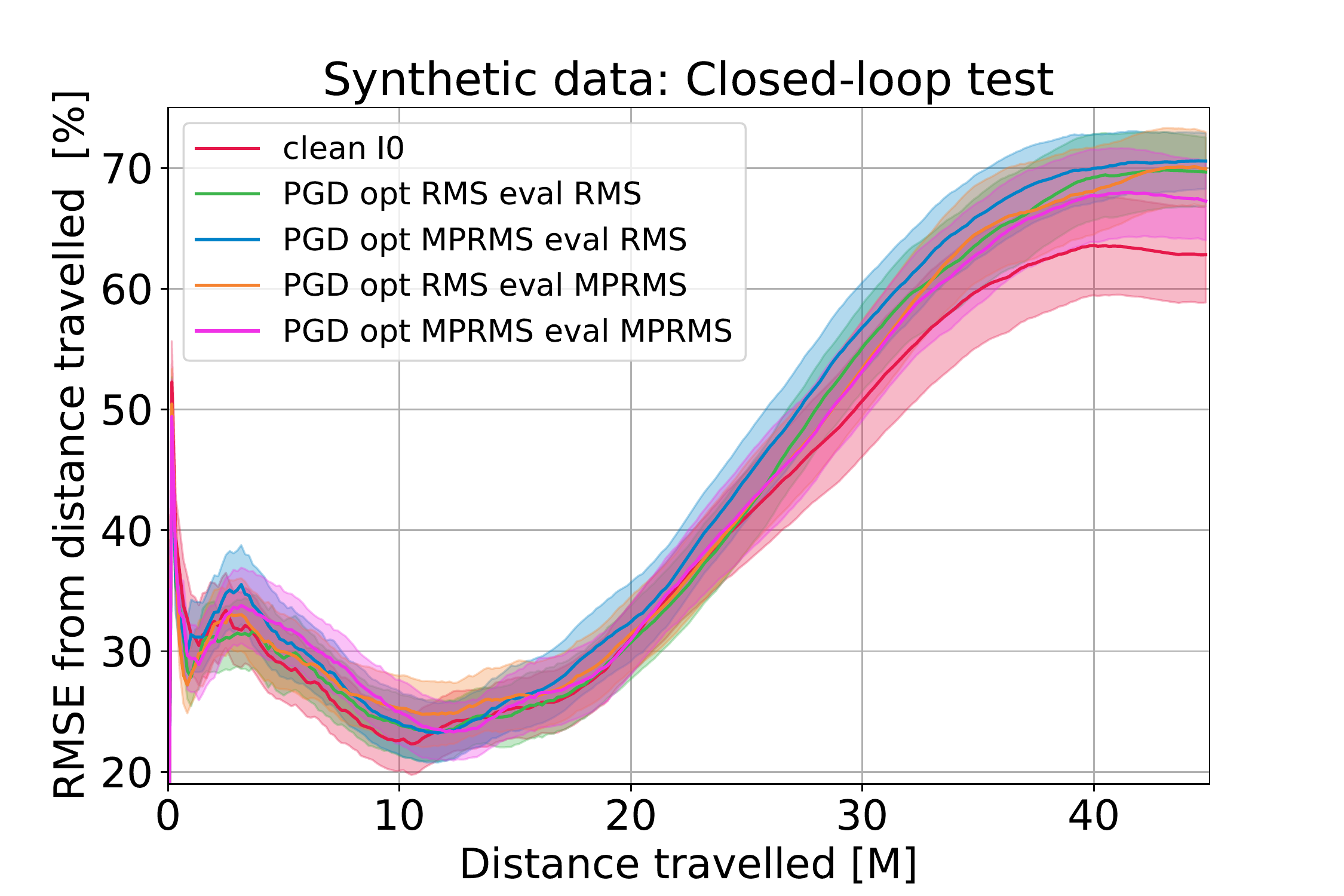}
    \includegraphics[width=0.483\linewidth]{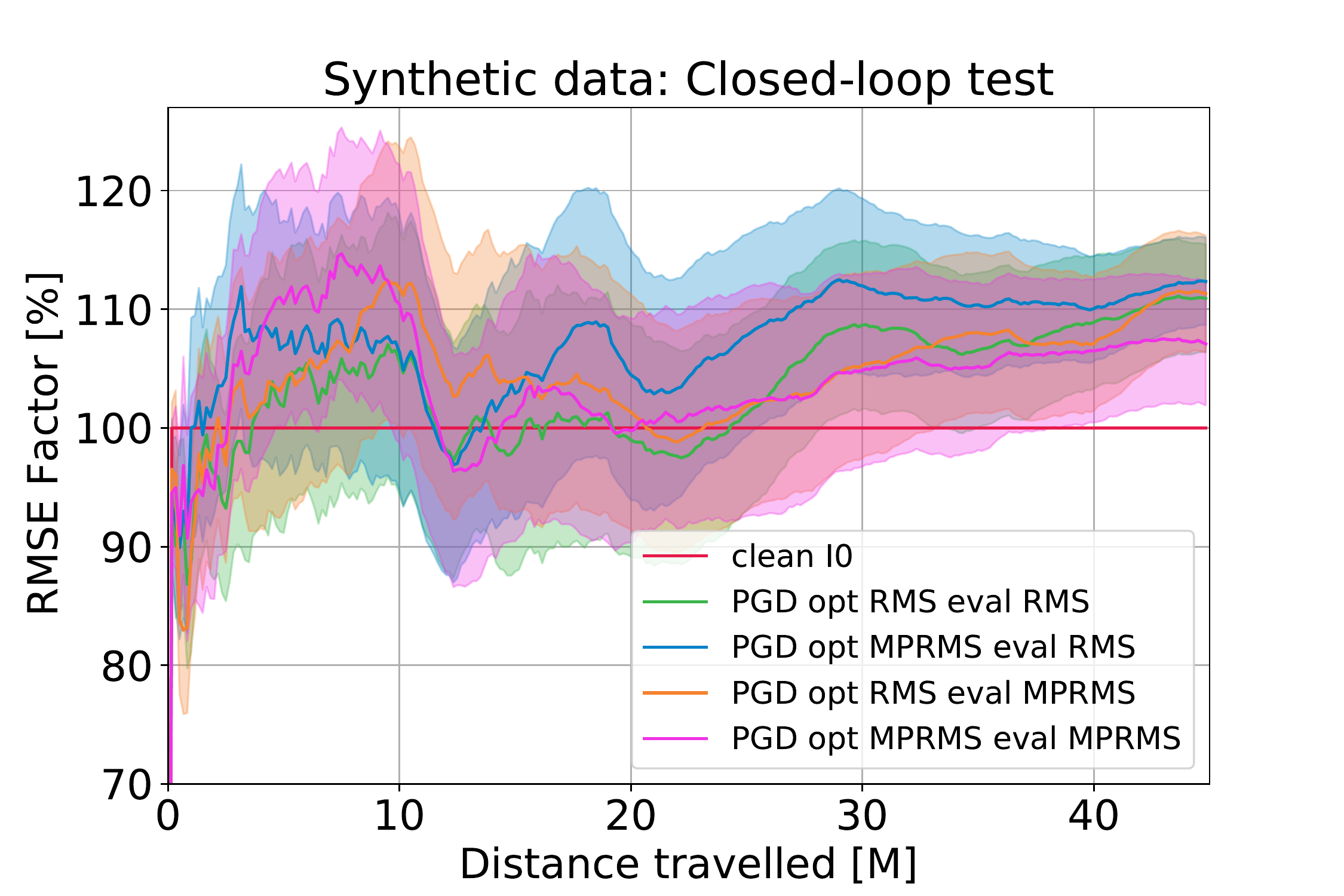}
 \caption{
   Accumulated deviation in distance travelled from the ground-truth over closed-loop trajectories of the synthetic dataset as a function of the trajectory length. We show a comparison of the deviation in distance travelled between our universal attacks and the clean baseline (left) as well as the ratio of the deviation compared to the clean results (right). We present mean and standard deviation over the trajectories for each trajectory length.
 }
\label{fig:synth_closedloop}
\end{figure*}

In \cref{fig:synth_closedloop} we show the closed-loop results on the synthetic dataset. Our universal attacks showed an increase in the generated deviation over the clean baseline, which, however, was not as substantial as before as the baseline's generated deviation is already quite significant. The best performing universal attack generated, after $45[m]$, a deviation of $71\%$, in distance travelled, which is a factor of $112\%$ from the clean baseline. Note that the adversarial patches that were optimized on relatively short trajectories are effective on longer trajectories in the closed-loop scheme, without any fine-tuning. We again see that the best performance is achieved for $\ell_{train}=\ell_{MPRMS}, \ell_{eval}=\ell_{RMS}$.
\subsubsection{Real data experiments}
\begin{figure*}
 \centering
    \includegraphics[width=0.483\linewidth]{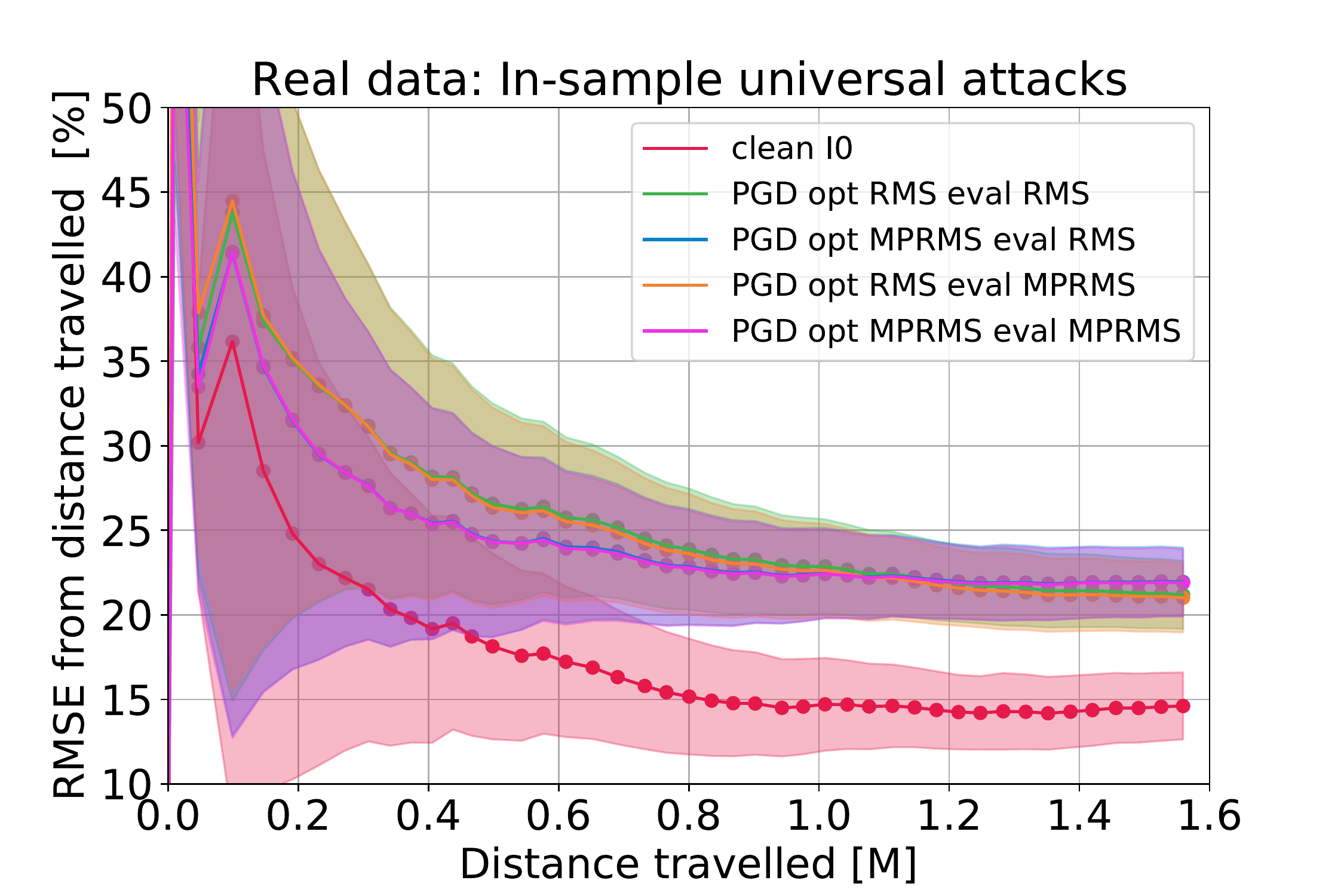}
    \includegraphics[width=0.483\linewidth]{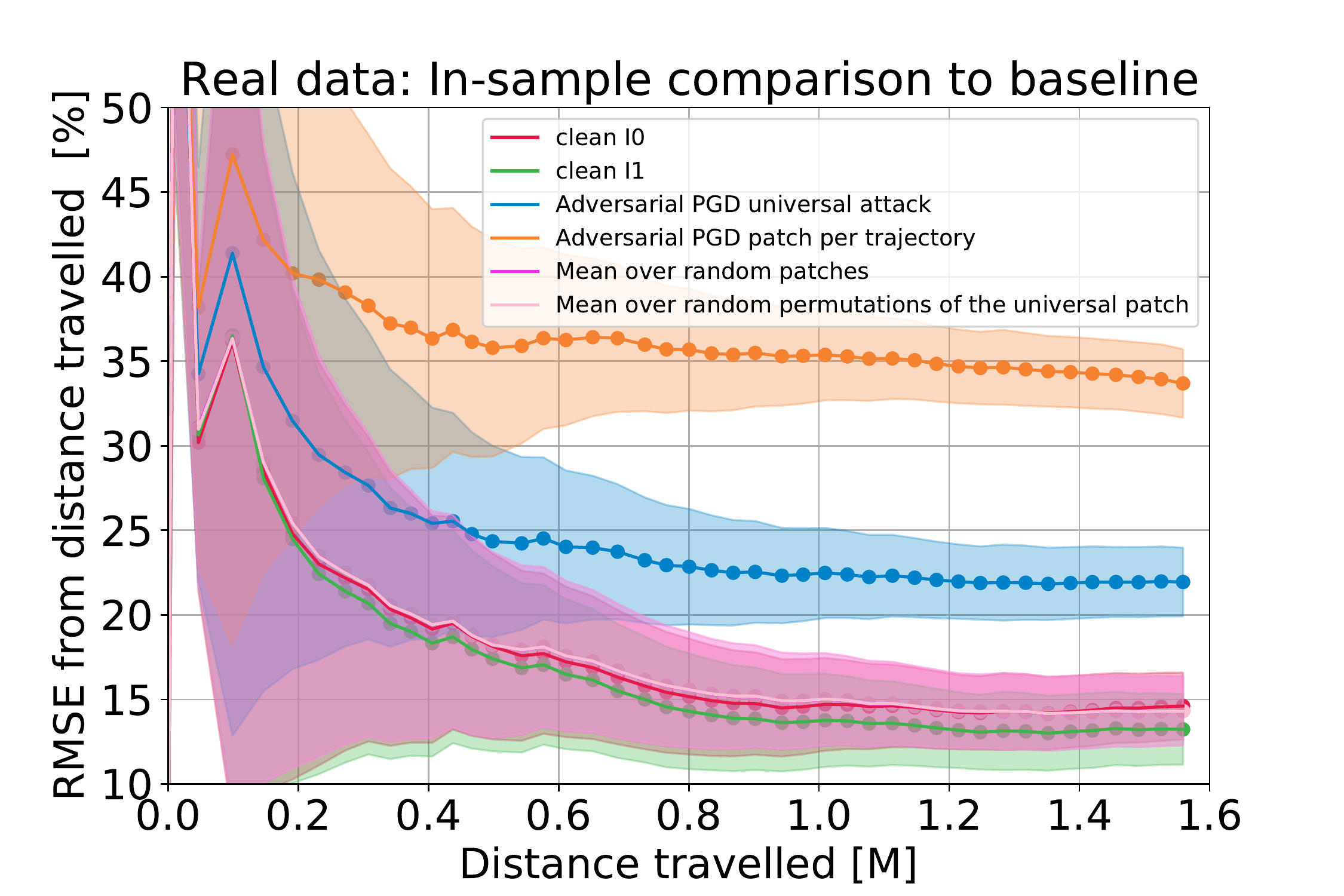}
 \caption{
 Accumulated deviation in distance travelled from the ground-truth trajectories on the real dataset as a function of the trajectory length. We show a comparison of our universal attacks trained on the entire dataset (left), and a comparison of our best performing universal and PGD attacks to the clean and random perturbation baselines (right). We present mean and standard deviation over the trajectories for each trajectory length.
 }
\label{fig:real_in_sample}
\end{figure*}

In \cref{fig:real_in_sample} we show the in-sample results on the real dataset. Similarly to the synthetic dataset, we see a substantial improvement for both our universal and PGD attacks over the clean $I^0$ baseline, while the clean $I^1$ and random baselines show a slight decrease. The best PGD attack generated, after $1.56[m]$, a deviation of $34\%$ in distance travelled, which is a factor of $231\%$ from the clean $I^0$ baseline. For the same configuration, the best universal attack generated a deviation of $22\%$ in distance travelled, which is a factor of $150\%$ from the clean $I^0$ baseline. The increase in the generated deviation is less significant compared to the synthetic dataset, partially due to the smaller patch size as in \cref{fig:synth_fov}.\\

\begin{figure*}
 \centering
    \includegraphics[width=0.483\linewidth]{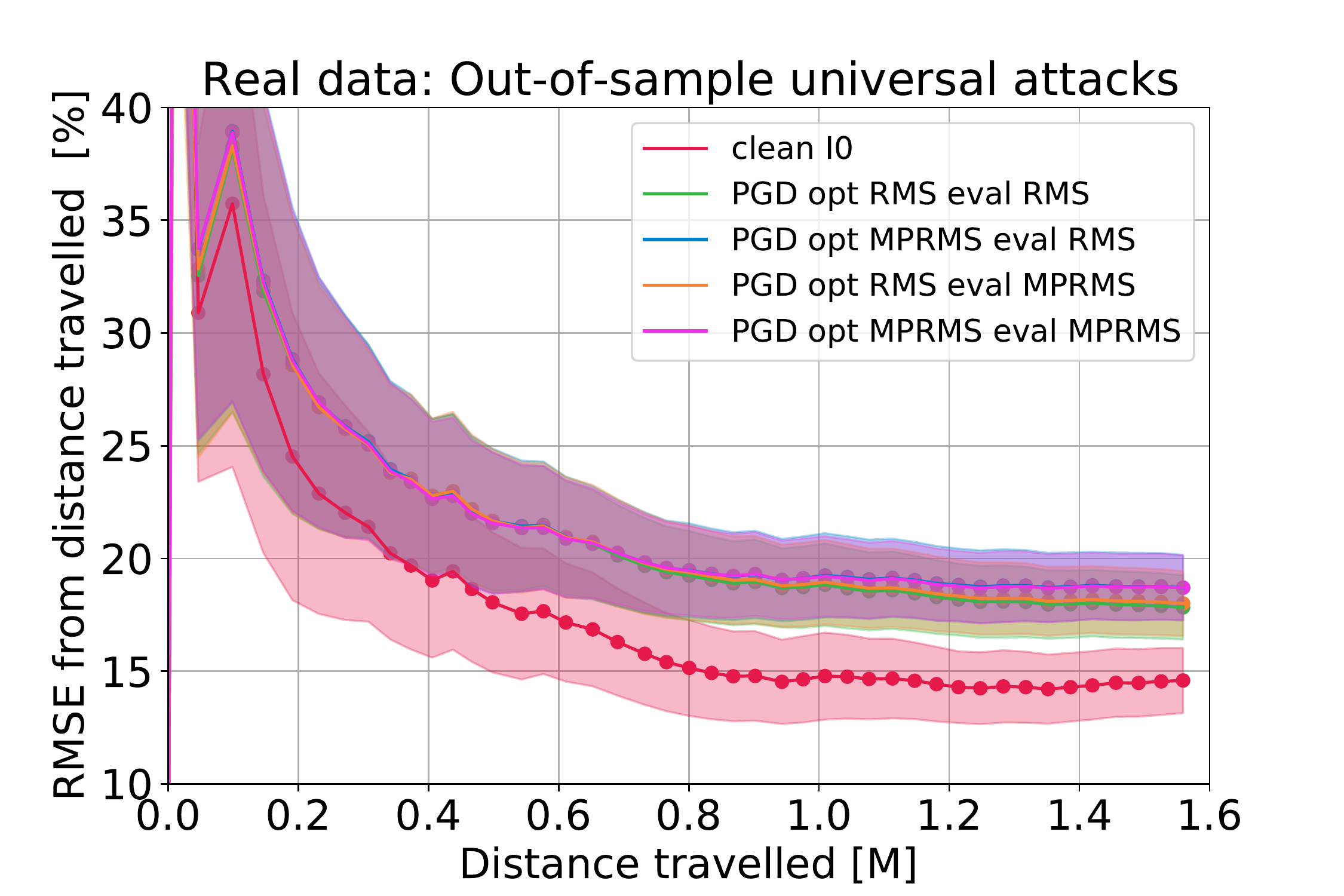}
    \includegraphics[width=0.483\linewidth]{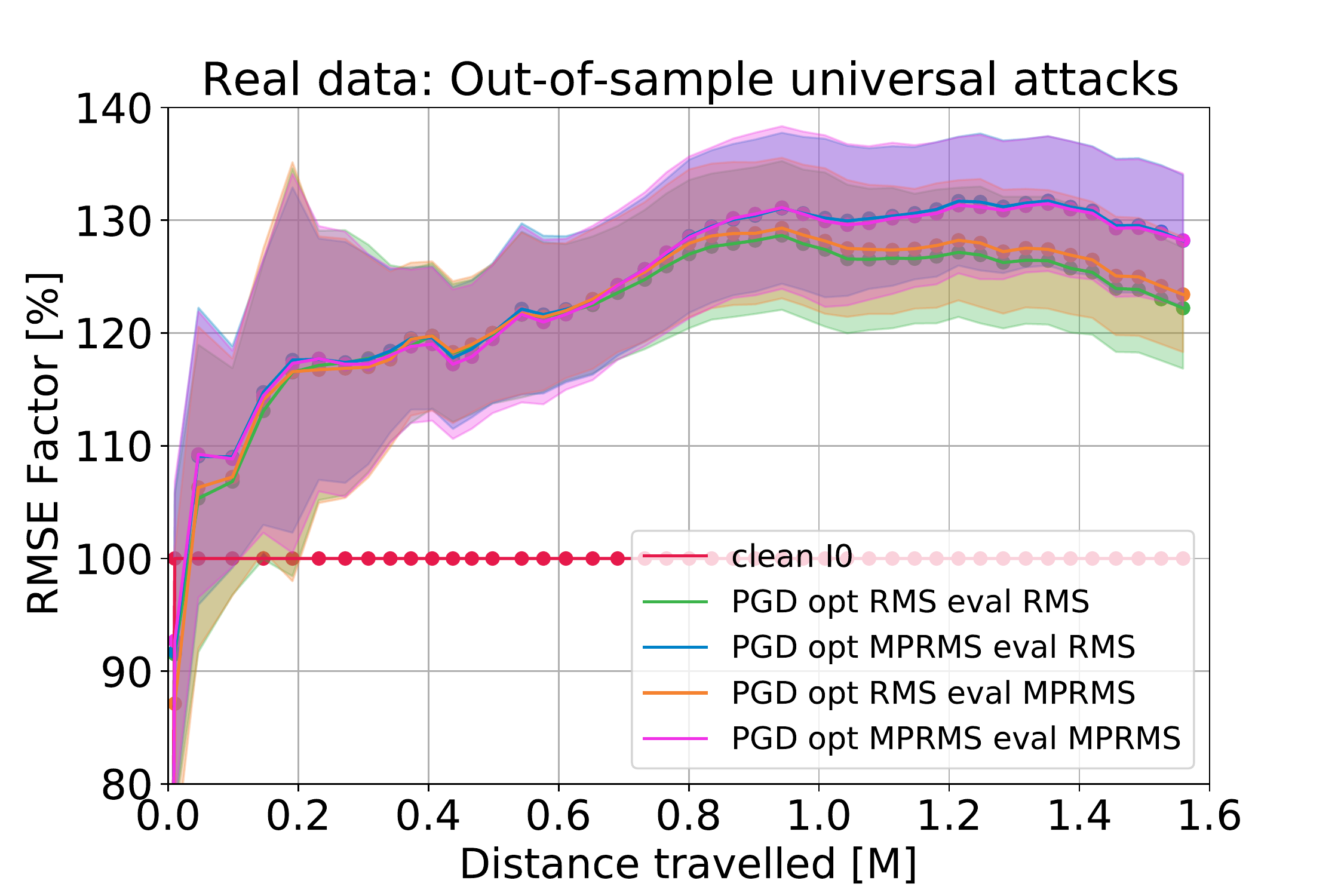}
 \caption{
  Accumulated deviation in distance travelled from ground-truth trajectories over out-of-sample cross-validation of the real dataset as a function of the trajectory length. We show a comparison of the deviation in distance travelled between our universal attacks and the clean baseline (left) as well as the ratio of the deviation compared to the clean results (right). We present mean and standard deviation over the trajectories for each trajectory length.
 }
\label{fig:real_oos}
\end{figure*}

In \cref{fig:real_oos} we show the out-of-sample results on the real dataset. Our universal attacks again showed an increase in the generated deviation over the clean baseline, with the best universal attack generating, after $1.56[m]$, a deviation of $19\%$ in distance travelled, which is a factor of $128\%$ from the clean $I^0$ baseline. The best performance is again achieved for the $\ell_{train}=\ell_{MPRMS}$ optimization criterion with negligible difference in the choice of $\ell_{eval}$.

\section{Conclusions}
\label{sec:conclusion}
This paper proposed a novel method for passive patch adversarial attacks on visual odometry-based navigation systems. We used homography of the adversarial patch to different viewpoints to understand how each perceives it and optimize the patch for entire trajectories. Furthermore, we limited the adversarial patch in the $\ell_{inf}$ and $\ell_{0}$ norms by taking into account the black and white albedo images of the patch and the FOV of the patch.

On the synthetic dataset, we showed that the proposed method could effectively force a given trajectory or set of trajectories to  deviate from their original path. For a patch FOV of $22.2\%$, our PGD attack generated, on a given trajectory, an average deviation, after $10[m]$, of $103\%$ in distance travelled, and given the entire trajectory dataset, our universal attack produced a single adversarial patch that generated an average deviation, after $10[m]$, of $80\%$ in distance travelled. Moreover, our universal attack generated, on out-of-sample data, a deviation, after $10[m]$, of $61\%$ in distance travelled and in a closed-loop setting generated an average deviation, after $45[m]$, of $71\%$ in distance travelled.

In addition, while less substantial, our results were replicated using the real dataset and a significantly smaller patch FOV of $8.8\%$. Nevertheless, when considering the effect with a larger patch FOV, we can expect a corresponding increase in the generated deviation as we saw in the synthetic dataset. For a given trajectory, our PGD attack generated an average deviation, after $1.56[m]$, of $34\%$ in distance travelled, and our universal attack generated an average deviation, after $1.56[m]$, of $22\%$ in distance travelled given the entire dataset, and on out-of-sample data generated an average deviation, after $1.56[m]$, of $19\%$ in distance travelled. 

We conclude that physical passive patch adversarial attacks on vision-based navigation systems could be used to harm systems in both simulated and real-world scenes. Furthermore, such attacks represents a severe security risk as they could potentially push an autonomous system onto a collision course with some object by simply inserting a pre-optimized patch into a scene. 

Our results were achieved using a predefined location for the adversarial patch. Optimizing the location of the adversarial patches may produce even more substantial results. For example, Ikram et al. (2022)\cite{ikram2022perceptual} showed that inserting a simple high-textured patch into specific locations in a scene produces false loop closures and thus degenerates state-of-the-art SLAM algorithms.

\paragraph{\textbf{Acknowledgements}}
This project was funded by GRAND/HOLDSTEIN drone technology competition, European Research Council (ERC) under the European Union’s Horizon 2020 research and innovation programme (grant agreement No. 863839), The Technion Hiroshi Fujiwara Cyber Security Research Center, and the Israel Cyber Directorate

\bibliographystyle{splncs}
\bibliography{egbib}

\clearpage
\appendix

\section{Adversarial attacks algorithms}
\label{sec:appendix}

We present algorithms for both our PGD (\cref{alg:attack}) and universal (\cref{alg:universal_attack}) attacks. In both cases we make use of the PGD adversarial attack scheme \cite{madry2017towards} to optimize a single adversarial patch. In each optimization step, we update the patch based on the gradient of the training criterion. Finally, we return the produced patch which maximized the evaluation criterion.

\begin{algorithm}
	\caption{PGD adversarial attack}
	\label{alg:attack}
	\hspace*{\algorithmicindent} \textbf{Input} $VO$: VO model\\
	\hspace*{\algorithmicindent} \textbf{Input} $A$: Adversarial patch perturbation\\
	\hspace*{\algorithmicindent} \textbf{Input} $(x, y)$: Trajectory to attack and it's ground truth motions \\
	\hspace*{\algorithmicindent} \textbf{Input} $(\ell_{train}, \ell_{eval})$: Train and evaluation loss functions\\
	\hspace*{\algorithmicindent} \textbf{Input} $\alpha$: Step size for the attack\\
	\begin{algorithmic}[H]
		\State $P \leftarrow \text{Uniform}(0, 1)$
		\State $P_\text{best} \leftarrow P$
		\State $\text{Loss}_\text{best} \leftarrow 0$
		\For{$k=1$ to $K$}
            \State \underline{\textbf{optimization step: }}
            \State $g \leftarrow \nabla_{P} \ell_{train}(VO(A(x,P)), y)$
            \State $P \leftarrow P + \alpha \cdot \text{sign}(g)$
            \State $P \leftarrow clip(P,0,1)$
            \State \underline{\textbf{evaluate patch: }}
            \State $\text{Loss} \leftarrow \ell_{eval}(VO(A(x,P)), y)$
	        \If{$\text{Loss} > \text{Loss}_\text{best}$}
                \State $P_\text{best} \leftarrow P$
                \State $\text{Loss}_\text{best} \leftarrow \text{Loss}$
	        \EndIf
    	\EndFor
		\State \textbf{return} $P_\text{best}$ 
	\end{algorithmic}
\end{algorithm}

\begin{algorithm}
	\caption{Universal PGD adversarial attack}
	\label{alg:universal_attack}
	\hspace*{\algorithmicindent} \textbf{Input} $VO$: VO model\\
	\hspace*{\algorithmicindent} \textbf{Input} $A$: Adversarial patch perturbation\\
	\hspace*{\algorithmicindent} \textbf{Input} $(X_{train}, Y_{train})$: Trajectories training dataset\\
	\hspace*{\algorithmicindent} \textbf{Input} $(X_{eval}, Y_{eval})$: Trajectories evaluation dataset\\
	\hspace*{\algorithmicindent} \textbf{Input} $(\ell_{train}, \ell_{eval})$: Training and evaluation loss functions\\
	\hspace*{\algorithmicindent} \textbf{Input} $(N_{train}, N_{eval})$: Number of training and evaluation trajectories\\
	\hspace*{\algorithmicindent} \textbf{Input} $\alpha$: Step size for the attack\\
	\begin{algorithmic}[H]
		\State $P \leftarrow \text{Uniform}(0, 1)$
		\State $P_\text{best} \leftarrow P$
		\State $\text{Loss}_\text{best} \leftarrow 0$
		\For{$k=1$ to $K$}
            \State \underline{\textbf{optimization step: }}
            \State $g \leftarrow 0$
	        \For{$i=1$ to $N_{train}$}
                \State $\hat{y}_{train, i} \leftarrow VO(A(x_{train, i},P))$
                \State $g \leftarrow g + \nabla_{P} \ell_{train}(\hat{y}_{train, i}, y_{train, i})$
	        \EndFor
            \State $P \leftarrow P + \alpha \cdot \text{sign}(g)$
            \State $P \leftarrow clip(P,0,1)$
            \State \underline{\textbf{evaluate patch: }}
            \State $\text{Loss} \leftarrow 0$
	        \For{$i=1$ to $N_{eval}$}
                \State $\hat{y}_{eval, i} \leftarrow VO(A(x_{eval, i},P))$
                \State $\text{Loss} \leftarrow \text{Loss} + \ell_{eval}(\hat{y}_{eval, i}, y_{eval, i})$
	        \EndFor
	        \If{$\text{Loss} > \text{Loss}_\text{best}$}
                \State $P_\text{best} \leftarrow P$
                \State $\text{Loss}_\text{best} \leftarrow \text{Loss}$
	        \EndIf
    	\EndFor
		\State \textbf{return} $P_\text{best}$ 
	\end{algorithmic}
\end{algorithm}

\section{Real data experiment specifics}
For the generation of the real dataset, in addition to the Mocap markers we make use of Aruco markers to produce the patch's coordinates in the camera system for each frame, which are then used for generating the patch's mask. In addition, the Aruco Markers, being printed on paper or some other material, provide an estimate for the albedo extremes of the printed patch on the same material. In each frame, we than make use of the detected patch to estimate its albedo limits. We calculate these limits by fitting the pixel histogram of the patch area to a Bivariate normal distribution. We account for the illumination variation within this area by multiplying our albedo images by the lightness channel of the HSL (hue, saturation, lightness) representation of the original image. As seen in \cref{fig:real_frame}, the black and white albedo images accordingly resemble the black and white pixels in the Aruco markers.

Throughout the data-set generation process, we discarded trajectories with incomplete camera pose or patch coordinates.

The trajectories initial positions formed an horizontal angle range of $[-8.5^{\circ},8.5^{\circ}]$ with respect to the target patch plane.

\end{document}